\documentclass[sigconf]{acmart}
\AtBeginDocument{%
  }

\setcopyright{acmlicensed}
\copyrightyear{2018}
\acmYear{2018}
\acmDOI{XXXXXXX.XXXXXXX}
\acmConference[Conference acronym 'XX]{Make sure to enter the correct
  conference title from your rights confirmation email}{June 03--05,
  2018}{Woodstock, NY}
\acmISBN{978-1-4503-XXXX-X/2018/06}

\usepackage{xspace}
\usepackage[export]{adjustbox}
\usepackage{multirow}
\usepackage{threeparttable}
\usepackage{color, colortbl}
\usepackage{makecell}
\usepackage{subcaption}
\usepackage{graphicx}

\usepackage[ruled,linesnumbered]{algorithm2e}
\usepackage{xcolor} 
\usepackage[most]{tcolorbox}

\begin{document}

\title{Multiple Consistent 2D-3D Mappings for Robust Zero-Shot 3D Visual Grounding}


\author{Yufei Yin}
\authornote{Both authors contributed equally to this research.}
\affiliation{%
  % \institution{School of Computer Science and Technology, Hangzhou Dianzi University}
  % \institution{Hangzhou Dianzi University}
  % \city{Hangzhou}
  % \state{Zhejiang}
  % \country{China}
  \institution{Hangzhou Dianzi University}
  \city{Hangzhou}
  \state{Zhejiang}
  \country{China}
}
\email{yinyf@hdu.edu.cn}

\author{Jie Zheng}
\authornotemark[1]
\affiliation{%
  % \institution{School of Computer Science and Technology}
  \institution{Hangzhou Dianzi University}
  \city{Hangzhou}
  \state{Zhejiang}
  \country{China}
}
\email{avengerjie@163.com}

% \author{Yufei Yin, Jie Zheng}
% \affiliation{
%   \city{Hangzhou Dianzi University}
%   \country{China}
% }
% \email{yinyf@hdu.edu.com}
% \email{avengerjie@163.com}

% \author{Jie Zheng}
% \affiliation{%
%   % \institution{School of Computer Science and Technology}
%   \city{Hangzhou Dianzi University}
%   \country{China}
% }
% \email{avengerjie@163.com}

\author{Qianke Meng}
\affiliation{%
  % \institution{School of Computer Science and Technology}
  \institution{Hangzhou Dianzi University}
  \city{Hangzhou}
  \state{Zhejiang}
  \country{China}
}
\email{mqk@hdu.edu.cn}

\author{Zhou Yu}
\authornote{Corresponding author.}
\affiliation{%
  % \institution{School of Computer Science and Technology}
  \institution{Hangzhou Dianzi University}
  \city{Hangzhou}
  \state{Zhejiang}
  \country{China}
}
\email{yuz@hdu.edu.cn}

\author{Minghao Chen}
\affiliation{%
  % \institution{School of Computer Science and Technology}
  \institution{Hangzhou Dianzi University}
  \city{Hangzhou}
  \state{Zhejiang}
  \country{China}
}
\email{chenminghao@hdu.edu.cn}

\author{Jiajun Ding}
\affiliation{%
  % \institution{School of Computer Science and Technology}
  \institution{Hangzhou Dianzi University}
  \city{Hangzhou}
  \state{Zhejiang}
  \country{China}
}
\email{djj@hdu.edu.cn}

\author{Min Tan}
\affiliation{%
  % \institution{School of Computer Science and Technology}
  \institution{Hangzhou Dianzi University}
  \city{Hangzhou}
  \state{Zhejiang}
  \country{China}
}
\email{tanmin@hdu.edu.cn}

\author{Yuling Xi}
\affiliation{%
  % \institution{School of Computer Science and Technology}
  \institution{Zhejiang University}
  \city{Hangzhou}
  \state{Zhejiang}
  \country{China}
}
\email{xiyuling@zju.edu.cn}

\author{Zhiwen Chen}
\affiliation{%
  % \institution{School of Computer Science and Technology}
  \institution{Alibaba Group}
  \city{Hangzhou}
  \state{Zhejiang}
  \country{China}
}
\email{zhiwen.czw@alibaba-inc.com}

\author{Chengfei Lv}
\affiliation{%
  % \institution{School of Computer Science and Technology}
  \institution{Alibaba Group}
  \city{Hangzhou}
  \state{Zhejiang}
  \country{China}
}
\email{chengfei.lcf@alibaba-inc.com}

\renewcommand{\shortauthors}{Trovato et al.}

\begin{abstract}
  Zero-shot 3D Visual Grounding (3DVG) is a critical capability for open-world embodied AI. However, existing methods are fundamentally bottlenecked by the poor quality of open-vocabulary 3D proposals, suffering from inaccurate categories and imprecise geometries, as well as the spatial redundancy of exhaustive multi-view reasoning. To address these challenges, we propose MCM-VG, a novel framework that achieves robust zero-shot 3DVG by explicitly establishing Multiple Consistent 2D-3D Mappings. Instead of passively relying on noisy 3D segments, MCM-VG enforces 2D-3D consistency across three fundamental dimensions to achieve precise target localization and reliable reasoning. First, a Semantic Alignment module corrects category mismatches via LLM-driven query parsing and coarse-to-fine 2D-3D matching. Second, an Instance Rectification module leverages VLM-guided 2D segmentations to reconstruct missing targets, back-projecting these reliable visual priors to establish accurate 3D geometries. Finally, to eliminate spatial redundancy, a Viewpoint Distillation module clusters 3D camera directions to extract optimal frames. By pairing these optimal RGB frames with Bird's Eye View maps into concise visual prompt sets, we formulate the final target disambiguation as a multiple-choice reasoning task for Vision-Language Models. 
  % Extensive evaluations on the prevalent ScanRefer dataset demonstrate that MCM-VG sets a new state-of-the-art for zero-shot 3D visual grounding. Remarkably, our framework achieves 62.0\% and 53.6\% in Acc@0.25 and Acc@0.5, outperforming previous zero-shot methods by substantial margins of 6.4\% and 4.0\%, respectively.
  Extensive evaluations on ScanRefer and Nr3D benchmarks demonstrate that MCM-VG sets a new state-of-the-art for zero-shot 3D visual grounding. Remarkably, it achieves 62.0\% and 53.6\% in Acc@0.25 and Acc@0.5 on ScanRefer, outperforming previous baselines by substantial margins of 6.4\% and 4.0\%.
\end{abstract}

%    <concept>
%        <concept_id>10010147.10010178.10010224.10010225.10010227</concept_id>
%        <concept_desc>Computing methodologies~Scene understanding</concept_desc>
%        <concept_significance>500</concept_significance>
%        </concept>
%  </ccs2012>

% \ccsdesc[500]{Computing methodologies~Scene understanding}

\begin{CCSXML}
<ccs2012>
   <concept>
       <concept_id>10010147.10010178.10010224.10010225.10010227</concept_id>
       <concept_desc>Computing methodologies~Scene understanding</concept_desc>
       <concept_significance>100</concept_significance>
       </concept>
 </ccs2012>
\end{CCSXML}

\ccsdesc[100]{Computing methodologies~Scene understanding}

\keywords{3D Visual Grounding, Zero-shot Scene Understanding, Cross-modal Alignment, Visual-Language Model}


\maketitle

\section{Introduction}

\begin{figure}
    \centering
    \includegraphics[width=0.98\linewidth]{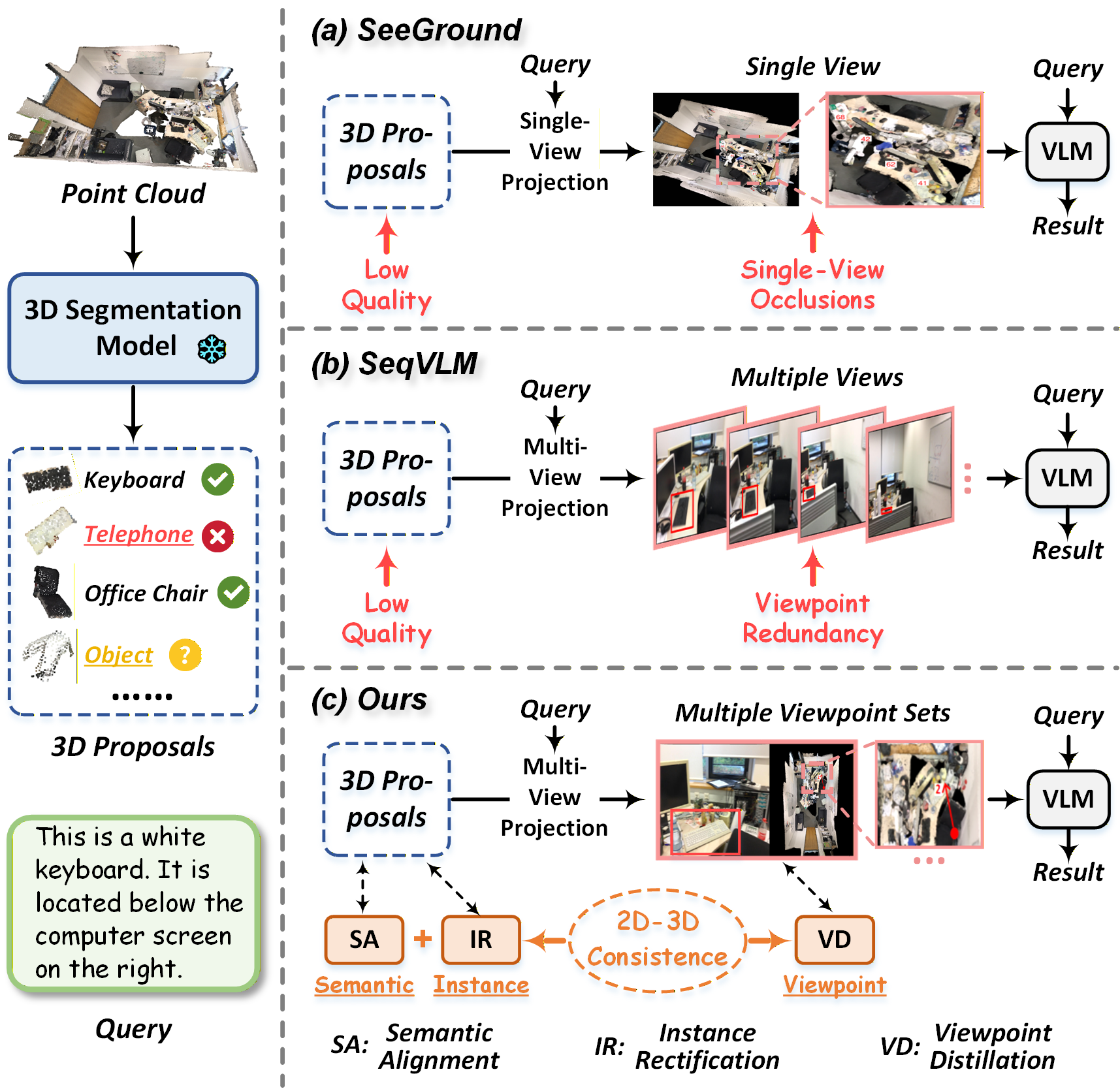}
    \vspace{-0.5em}
    \caption{Comparisons with zero-shot 3DVG methods. Compared with SeeGround \cite{li2025seeground} and  SeqVLM \cite{lin2025seqvlm} that passively rely on noisy 3D proposals or perform exhaustive reasoning over redundant multi-view projections, MCM-VG explicitly enforces rigorous 2D-3D consistency at the semantic, instance, and viewpoint levels, thereby improving both grounding accuracy and reasoning reliability.}
    \label{fig:intro}
    \vspace{-0.3cm}
\end{figure}

3D Visual Grounding (3DVG) is a fundamental capability for embodied AI and intelligent robotics, enabling agents to localize specific objects within a 3D environment based on natural language queries. Given a 3D scene, typically represented as a point cloud, and a free-form textual description, this task requires deep comprehension of both complex linguistic semantics and fine-grained 3D spatial structures to achieve precise cross-modal alignment.

To date, the predominant paradigm in 3DVG relies heavily on fully supervised learning \cite{chen2020scanrefer, huang2021text, yang2021sat, huang2022multi, luo20223d, wu2023eda, zhang2024vision, qian2024multi}. While these methods achieve impressive performance, they suffer from two critical limitations. First, they are notoriously data-hungry, requiring vast amounts of meticulously annotated 3D-text pairs, which are resource-intensive to acquire and difficult to scale. Second, their performance is strictly bounded by a predefined vocabulary, rendering them brittle and sub-optimal in open-world scenarios where novel, previously unseen objects must be grounded.

To circumvent these bottlenecks, zero-shot and open-vocabulary 3DVG methods \cite{peng2023openscene, yang2024llm, yuan2024visual, xu2024vlm, li2025seeground, lin2025seqvlm} have emerged as promising alternatives. By harnessing the formidable generalization capabilities of pre-trained Large Language Models (LLMs) \cite{touvron2023llama,guo2025deepseek,yang2025qwen3} and Vision-Language Models (VLMs) \cite{singh2025openai,bytedance_doubao16,bai2025qwen3}, these approaches eliminate the need for task-specific 3D-text annotations. A typical pipeline follows a two-stage process: first, an off-the-shelf 3D segmentation model extracts class-agnostic or open-vocabulary 3D proposals from the point cloud; second, these 3D proposals are transformed into textual or visual representations and fed into LLMs/VLMs for spatial reasoning and target matching.

Despite their potential, existing zero-shot 3DVG methods face three major limitations. First, pure LLM-based approaches \cite{yang2024llm, yuan2024visual} typically serialize 3D scene graphs or proposals into structured text. This not only incurs substantial token overhead but also inevitably discards rich visual cues such as color, texture, and fine-grained geometry. Second, while VLM-based methods \cite{li2025seeground, lin2025seqvlm} attempt to preserve visual details via rendered point clouds or projected video frames, they often struggle with visual quality and viewpoint redundancy. SeeGround \cite{li2025seeground} (Fig. ~\ref{fig:intro} (a)) employs query-aligned dynamic rendering, but rendered point clouds suffer from artifacts and single-view occlusions. Conversely, SeqVLM \cite{lin2025seqvlm} (Fig. ~\ref{fig:intro} (b)) projects 3D proposals onto multi-view frames for iterative reasoning; however, this introduces severe viewpoint redundancy and each 2D view only provides partial local spatial evidence. Third, and most crucially, both paradigms are fundamentally bottlenecked by the quality of the initial 3D proposals. Inaccurate category predictions lead to fatal semantic mismatches, and imprecise 3D bounding boxes severely degrade cross-view reasoning and final grounding accuracy.

In this paper, we propose MCM-VG, a novel framework that leverages \textbf{M}ultiple \textbf{C}onsistent 2D-3D \textbf{M}appings to achieve robust zero-shot 3D \textbf{V}isual \textbf{G}rounding. As illustrated in Fig.~\ref{fig:intro} (c), rather than passively relying on noisy 3D proposals or performing exhaustive reasoning over redundant multi-view projections, MCM-VG explicitly enforces rigorous 2D-3D consistency at the semantic, instance, and viewpoint levels, thereby improving both grounding accuracy and reasoning reliability.
Specifically, MCM-VG operates in three distinct stages. First, we introduce a semantic alignment module to enforce semantic consistency between 2D observations and 3D proposals. By combining LLM-driven query parsing with a coarse-to-fine 2D-3D matching mechanism, we effectively correct the noisy category predictions inherent in initial 3D proposals. Second, we propose an instance rectification module to guarantee instance consistency and precise localization. Acting as a robust fallback mechanism when initial 3D proposals completely fail, it leverages VLM-guided 2D segmentations to reconstruct missing targets, back-projecting these reliable visual priors to independently establish an accurate target set. Finally, we design a viewpoint distillation module to ensure viewpoint consistency and minimize VLM reasoning redundancy. By clustering 3D camera directions to extract optimal frames, we explicitly pair them with Bird's Eye View maps, distilling concise visual prompt sets for reliable 3D reasoning and grounding.

Our contributions are summarized as follows:
% \vspace{-0.4cm}
\begin{itemize}
    \item[$\bullet$] We propose MCM-VG, a novel and robust framework for zero-shot 3D visual grounding. By explicitly establishing multiple consistent 2D-3D mappings, our approach effectively bridges the gap between 2D visual priors and 3D geometries, overcoming the fundamental bottlenecks of noisy 3D proposals and redundant multi-view reasoning.
    \item[$\bullet$] We design three tailored modules for comprehensive 2D-3D mappings: a semantic alignment module to correct semantic mismatches, an instance rectification module to rectify instance localization, and a viewpoint distillation module to distill the optimal viewpoint for final VLM reasoning.
    % \item[$\bullet$] Extensive experiments on the widely used ScanRefer and Nr3D benchmarks demonstrate that MCM-VG achieves new state-of-the-art performance.
    \item[$\bullet$] We conduct extensive experiments on the widely used ScanRefer \cite{chen2020scanrefer} and Nr3D \cite{achlioptas2020referit3d} benchmarks, achieving new state-of-the-art performance and demonstrating the effectiveness of our proposed MCM-VG.
    % , setting a new state-of-the-art performance for zero-shot 3D visual grounding.
\end{itemize}

\vspace{-0.2cm}
\section{Related Work}
\subsection{Supervised 3D Visual Grounding}

Existing supervised 3D visual grounding methods mainly follow two paradigms: two-stage pipelines \cite{chen2020scanrefer,achlioptas2020referit3d,huang2021text,yang2021sat,huang2022multi,luo20223d,guo2023viewrefer,shi2024aware,wang2025augrefer}, which first generate candidate objects and then match them with the language query, and end-to-end pipelines \cite{wu2023eda,zhang2024vision,qian2024multi,geng2025pseudo,guo2025text}, which directly predict the target location from joint 3D-text representations. Recent advances in supervised 3DVG have focused mainly on handling viewpoint ambiguity and improving efficiency. For viewpoint reasoning, Pseudo-EV \cite{geng2025pseudo} formulates grounding as a viewpoint-aware reasoning process with LLM-generated pseudo embodied viewpoint labels. 
For efficiency, TSP3D \cite{guo2025text} introduces a single-stage sparse-convolutional framework, where text-guided voxel pruning and completion-based addition enable efficient multimodal interaction while maintaining strong localization performance.
Despite their strong benchmark performance, these supervised methods still rely heavily on large-scale annotations and closed-set categories, thereby limiting their broader generalization to unseen objects and open-world scenes. This limitation motivates recent research on zero-shot 3D visual grounding.

\vspace{-0.2cm}

\subsection{Zero-Shot 3D Visual Grounding}
To overcome the data dependency and limited generalization of fully-supervised methods, recent studies have explored zero-shot 3D visual grounding via large pre-trained models \cite{peng2023openscene, yang2024llm, yuan2024visual, xu2024vlm, li2025seeground, lin2025seqvlm}. These approaches generally fall into two categories: LLM-based and VLM-based. LLM-based methods \cite{yang2024llm, yuan2024visual} typically parse queries into structured programs for geometric reasoning in 3D space. While effective at semantic decomposition, they inherently struggle to exploit fine-grained visual cues like color and texture.
To bridge this gap, recent VLM-based methods \cite{li2025seeground, lin2025seqvlm} incorporate 2D visual evidence into the 3D grounding pipeline. 
SeeGround~\cite{li2025seeground} enhances grounding with query-relevant rendered views and cross-view alignment, while SeqVLM~\cite{lin2025seqvlm} further projects 3D proposals onto multi-view real images and performs proposal-guided reasoning over image sequences, improving the use of contextual and appearance information. These methods demonstrate the importance of incorporating 2D observations for 3DVG, especially when resolving ambiguous language descriptions or occluded objects.
% For instance, SeeGround \cite{li2025seeground} utilizes query-aligned rendered views, while SeqVLM \cite{lin2025seqvlm} projects 3D proposals onto multi-view video frames for sequential reasoning. Although these paradigms highlight the critical role of 2D observations in resolving semantic ambiguities, simply introducing more views inevitably leads to severe spatial redundancy. 

Motivated by this, our method moves beyond naively aggregating multi-view projections. Instead, we explicitly establish multiple consistent 2D-3D mappings that seamlessly couple 3D proposals with 2D visual evidence. This design systematically mitigates cross-modal inconsistencies and suppresses redundant viewpoints, providing VLMs with both fine-grained local details and global spatial awareness for robust zero-shot 3D grounding.

\begin{figure*}
    \centering
    \includegraphics[width=1.00\linewidth]{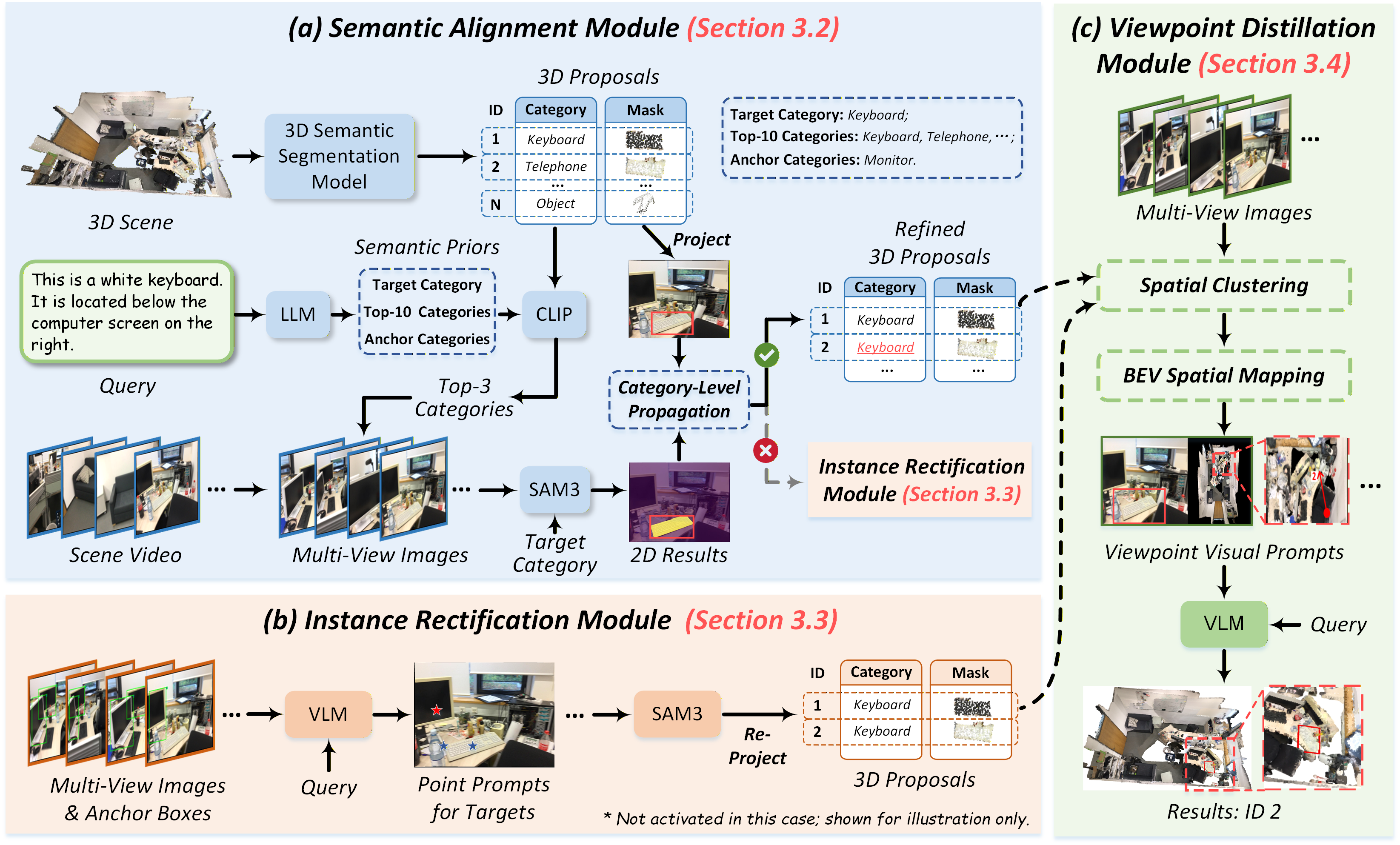}

    % \caption{Overview of the MCM-VG pipeline. First, the \textbf{Semantic Alignment} module parses $\mathcal{Q}$ into target and anchor priors via a LLM, and projects initial 3D proposals onto 2D planes to explicitly correct category mismatches through coarse-to-fine matching with 2D foundation models. Second, for geometrically inaccurate proposals, the \textbf{Instance Rectification} module leverages multi-modal foundation models to refine 2D segmentations, which are then seamlessly back-projected to augment and rectify the 3D localizations. Finally, to eliminate viewpoint redundancy, the \textbf{Viewpoint Distillation} module clusters camera viewing directions in 3D space to extract optimal frames. By explicitly pairing these RGB frames with Bird's Eye View (BEV) maps, it distills viewpoint visual prompt set. Ultimately, the visual prompt set and the query are fed into a Vision-Language Model (VLM) to deduce the final grounded box via multiple-choice spatial reasoning.}
    \vspace{-0.1cm}
    % \caption{Overview of the proposed MCM-VG pipeline for zero-shot 3D visual grounding.}
    \caption{Overview of the MCM-VG pipeline for zero-shot 3D visual grounding. MCM-VG contains three modules for comprehensive 2D-3D mappings: (a) Semantic Alignment Module to correct semantic mismatches, (b) Instance Rectification Module to rectify instance localization, and (c) Viewpoint Distillation Module to distill the optimal viewpoint for final VLM reasoning.}
    \vspace{-0.1cm}
    
    \label{fig:framework}
    % \vspace{-1.1em}
\end{figure*}

\section{Method}
% In this section, we present the methodology for our approach to 3D visual grounding, which integrates three key modules: Semantic Alignment Module, Instance Rectification Module, and Viewpoint Selection Module.

% \subsection{Task Definition}
% The task of 3DVG involves localizing a target object within a 3D scene based on a natural language query. The goal is to generate an accurate 3D bounding box that identifies the target object’s position and dimensions within the scene. Formally, this process can be expressed as:
% \begin{equation}
%     \text{bbox} = \text{3DVG}(S, Q)
% \end{equation}
% where \(S\) denotes the 3D scene, represented by a colored point cloud $\mathbf{P} \in \mathbb{R}^{N \times 6}$ and \(Q\) represents the natural language query. Our task focuses on zero-shot 3DVG that localizes the \(O^*\) in 3D space without requiring scene-specific training or fine-tuning. The image list can be obtained through various sensors including RGB-D sensors with structured light scanners 

\subsection{Overview}

Given a point cloud $\mathcal{P}\in \mathbb{R}^{N \times 6}$ and $M$ multi-view images $\mathcal{I}$ sampled from the scene video with their associated camera poses, along with a natural language query $\mathcal{Q}$, zero-shot 3D visual grounding aims to predict a 3D bounding box $\mathcal{B}^*$ for the target object without relying on task-specific annotations. To achieve this, we propose \textbf{MCM-VG}, a novel framework that explicitly enforces cross-modal alignment across semantics, instances, and viewpoints via \textbf{M}ultiple \textbf{C}onsistent 2D-3D \textbf{M}appings. As illustrated in Fig.~\ref{fig:framework}, our pipeline operates in three sequential stages. First, the \textit{Semantic Alignment} module parses $\mathcal{Q}$ into target and anchor priors via a LLM, and projects initial 3D proposals onto 2D planes to explicitly correct category mismatches through coarse-to-fine matching with 2D foundation models. Second, for geometrically inaccurate proposals, the \textit{Instance Rectification} module leverages multi-modal foundation models to refine 2D segmentations, which are then seamlessly back-projected to augment and rectify the 3D localizations. Finally, to eliminate viewpoint redundancy, the \textit{Viewpoint Distillation} module clusters camera viewing directions in 3D space to extract optimal frames. By explicitly pairing these RGB frames with Bird's Eye View maps, it distills a concise visual prompt set $\mathcal{V}$. Ultimately, $\mathcal{V}$ and the query $\mathcal{Q}$ are fed into a Vision-Language Model to deduce the final grounded box $\mathcal{B}^*$ via multiple-choice spatial reasoning.

\subsection{Semantic Alignment Module}
Off-the-shelf 3D segmentation models frequently struggle with open-vocabulary recognition, leading to severe semantic mismatches in zero-shot scenarios. To bridge this gap, the Semantic Alignment module explicitly enforces semantic consistency among the textual query, 2D visual priors, and 3D proposals.

Specifically, following previous work \cite{lin2025seqvlm}, we first process the point cloud scene $\mathcal{P}$ to generate an initial object proposal set $\mathcal{O}$ using a 3D instance segmentation network $\Phi_{\text{3D}}$ \cite{schult2023mask3d}:
\begin{equation}
    \mathcal{O} = \left\{ (M_k, B_k, C_k) \in \Phi_{\text{3D}}(\mathcal{P})\right\}_{k=1}^{K},
\end{equation}
where $M_k$, $B_k$, and $C_k$ denote the 3D instance mask, the 3D bounding box, and the predicted category, respectively. 

However, due to the inherent vocabulary limitations of pre-trained 3D networks, the predicted category $C_k$ frequently suffers from severe inaccuracies or semantic ambiguities (e.g., assigning a generic label like ``object'' to unobserved items). To address this bottleneck, we explicitly introduce cross-modal 2D-3D semantic consistency to rectify these erroneous categorical priors.

To achieve this, we prompt a Large Language Model to parse the free-form query $\mathcal{Q}$, extracting the target object category and the contextual anchor categories, along with a set of top-10 potential target category proposals $\mathcal{C}_{\text{top-10}}$. To ground these semantic priors visually, we project the 3D proposals whose predicted categories fall within $\mathcal{C}_{\text{top-10}}$ onto the multi-view 2D frames. 
We then utilize the text and visual encoders of CLIP~\cite{radford2021learning} to compute their respective embeddings $E_t$ and $E_v^{(k)}$, respectively, where $k$ represents the $k$-th 3D proposal.
% We then utilize the encoders of CLIP~\cite{radford2021learning} to compute their respective embeddings:
% \begin{align}
%     E_t &= f_{\text{text}}(T_{\text{target}}), \\
%     E_v^{(k)} &= f_{\text{visual}}(v_k), \quad \text{with} \ \ v_k = f_{\text{proj}}(M_k, \mathcal{I}),
% \end{align}
% where $f_{\text{text}}(\cdot)$ and $f_{\text{visual}}(\cdot)$ represent the text and visual encoders, respectively. Here, $f_{\text{proj}}(\cdot)$ acts as the mapping function, and $v_k$ denotes the 2D visual crop obtained by projecting the 3D mask $M_k$ onto the multi-view frames $\mathcal{I}$. 
To measure the cross-modal semantic alignment, the cosine similarity $\zeta_k$ between the target text embedding and the projected visual embedding is computed as:
\begin{equation}
    \zeta_k = \frac{E_t \cdot E_v^{(k)}}{\| E_t \| \| E_v^{(k)} \|}.
\end{equation}
Based on this similarity metric, we identify the top-three matching 3D proposals and extract their initial categories to form a refined potential target category set $\mathcal{C}_{\text{top-3}}$. To further conduct fine-grained semantic verification, we focus exclusively on the subset of 3D proposals whose predicted categories belong to $\mathcal{C}_{\text{top-3}}$. For each qualifying proposal $k$, we first extract its corresponding subset of visible 2D multi-view frames, denoted as $\mathcal{I}_k \subset \mathcal{I}$. Rather than relying on potentially noisy predicted categories, we explicitly leverage the LLM-parsed primary target category $T_{\text{target}}$ as the textual prompt for an advanced 2D segmentation model $\Psi_{\text{SEG}}$. For each visible frame $I_f \in \mathcal{I}_k$, the model generates a set of 2D bounding boxes along with their respective confidence scores:
\begin{equation}
    \hat{\mathcal{B}}_{f} = \Psi_{\text{SEG}}(I_f, T_{\text{target}}) = \left\{ (\hat{b}_{j}, \hat{s}_{j}) \right\}_{j=1}^{m_f},
\end{equation}
where $m_f$ represents the number of detected 2D boxes in frame $I_f$. 

For each qualifying 3D proposal $k$ with its 3D bounding box $B_k$, we project it onto the 2D image plane to obtain the projected 2D box $b_{k,f}$. To evaluate the spatial alignment between the 3D proposal and the 2D visual priors, we identify the best-matching 2D box in frame $f$ by calculating the maximum Intersection over Union (IoU):
\begin{equation}
    j^* = \arg\max_{j=1, \dots, m_f} \left( \text{IoU}(b_{k,f}, \hat{b}_{j}) \right).
\end{equation}
Let $\hat{b}_{j^*}$ and $\hat{s}_{j^*}$ denote this best-matched 2D box and its confidence score, respectively. The spatial matching score $\eta_{k}$ for the $k$-th 3D proposal is then aggregated across all $N_k$ visible multi-view frames by weighting the maximal IoU with the confidence score:
\begin{equation}
    \eta_{k} = \frac{1}{N_k}\sum_{f=1}^{N_k} \left( \text{IoU}(b_{k,f}, \hat{b}_{j^*}) \times \hat{s}_{j^*} \right).
\end{equation}

To ensure rigorous semantic consistency while mitigating the impact of false negatives in 2D predictions (e.g., low recall), we dynamically filter and update the 3D proposals using a category-level propagation strategy. Specifically, if the matching score $\eta_{k}$ of a proposal exceeds a predefined threshold $\gamma$, we consider its 3D geometry and the 2D target semantics to be reliably aligned. We collect the original 3D categories of all such successfully matched proposals to form a verified category set $\mathcal{C}_{\text{valid}}$.
% \begin{equation}
%     \mathcal{C}_{\text{valid}} = \left\{ C_i \mid \eta_i \geq \gamma \right\}.
% \end{equation}
To salvage valid targets that might have been missed by the 2D segmentation model, we additionally retain any unmatched proposal whose original 3D category $C_k$ belongs to $\mathcal{C}_{\text{valid}}$. For the semantic update, proposals that are directly matched ($\eta_{k} \geq \gamma$) are explicitly rectified to the definitive target category $T_{\text{target}}$. Meanwhile, to avoid injecting unverified semantics, the salvaged proposals ($\eta_{k} < \gamma$) conservatively retain their original 3D category $C_k$. 
% Formally, the rectified category $\tilde{C}_k$ for a retained proposal is defined as:
% \begin{equation}
%     \tilde{C}_k = 
%     \begin{cases} 
%         T_{\text{target}}, & \text{if } \eta_{k} \geq \gamma \\ 
%         C_k, & \text{otherwise.} 
%     \end{cases}
% \end{equation}
All remaining proposals whose original categories fall outside $\mathcal{C}_{\text{valid}}$ are discarded. Consequently, the initial object proposal set is successfully pruned and updated to form the semantically aligned set $\mathcal{O}'$.

\subsection{Instance Rectification Module}

As output by the Semantic Alignment module (Sec. 3.2), there are critical scenarios where \textit{all} candidate proposals fail to meet the matching threshold (i.e., $\max_k \eta_{\text{max}}^{(k)} < \gamma$). This severe failure indicates that the initial 3D segmentation has completely missed the target, typically bottlenecked by fundamental geometric inaccuracies rather than pure semantic errors. To rescue the localization process and reconstruct the missing 3D geometries, we explicitly shift to trusting 2D visual priors.

First, we identify the subset of 3D proposals $\mathcal{U}$ belonging to the top-3 candidate categories. For each proposal $k \in \mathcal{U}$, we extract its corresponding subset of visible 2D multi-view frames $\mathcal{I}_k \subset \mathcal{I}$.
% \begin{equation}
%     \mathcal{U} = \left\{ k \mid \eta_{\text{max}}^{(k)} < \gamma, \; C_k \in \mathcal{C}_{\text{top-3}} \right\}.
% \end{equation}

While Large Vision-Language Models (VLMs) trained on massive multi-modal data possess formidable zero-shot semantic localization, they inherently struggle with pixel-level dense boundary prediction. Conversely, 2D segmentation models excel at pixel-perfect segmentation but critically lack open-vocabulary semantic comprehension. To synergize their strengths, we design a task-specific prompt encapsulating the language query $\mathcal{Q}$. Furthermore, to provide the VLM with explicit spatial references, we project the 3D proposals associated with the parsed anchor categories $\mathcal{C}_{\text{anchor}}$ onto the extracted 2D frames $\mathcal{I}_k$, explicitly overlaying them with bounding boxes to serve as contextual visual cues. By feeding the language query alongside these anchor-annotated frames into a VLM (e.g., Gemini~\cite{team2023gemini}), the model leverages its rich cross-modal priors to output a set of precise 2D point coordinates $\hat{P}_{k,f}$ for the target object in the $f$-th frame.
% \begin{equation}
%     \hat{P}_{k,f} = \Psi_{\text{VLM}}(I_f, \mathcal{Q}), \quad \forall I_f \in \mathcal{I}_k,
% \end{equation}
% where $\Psi_{\text{VLM}}(\cdot)$ denotes the VLM inference function. 
To obtain fine-grained instance masks, we subsequently feed these multi-point coordinates $\hat{P}_{k,f}$ as explicit geometric prompts into the 2D segmentation model $\Psi_{\text{SEG}}$. This allows the model to segment the target precisely, producing robust 2D segmentations $\hat{B}_{k,f}$.
% \begin{equation}
%     \hat{B}_{k,f} = \text{SAM3}(I_f, \hat{P}_{k,f}).
% \end{equation}

Finally, we back-project these refined 2D segmentations from the multi-view frames into the 3D space. Through a robust integration process that includes cross-view frustum intersection and spatial denoising, we generate highly accurate, rectified 3D masks and bounding boxes $(M^*_k, B^*_k)$:
\begin{equation}
    M^*_k, B^*_k = f_{\text{back-proj}} \left( \left\{ \hat{B}_{k,f} \right\}_{I_f \in \mathcal{I}_k}, \mathcal{P} \right),
\end{equation}
where $f_{\text{back-proj}}(\cdot)$ represents the depth-aware 2D-to-3D mapping and denoising function based on point cloud $\mathcal{P}$. Since the initial 3D proposals were deemed completely unreliable, we discard the flawed geometric priors. Instead, these successfully recovered instances directly constitute the newly established target set $\mathcal{O}'$:
\begin{equation}
    \mathcal{O}' = \left\{ (M^*_k, B^*_k, T_{\text{target}}) \right\}_{k \in \mathcal{U}}.
\end{equation}
Through this synergistic project-and-rectify fallback mechanism, MCM-VG effectively reconstructs the missing targets directly from 2D visual priors, guaranteeing robust 2D-3D instance consistency even when off-the-shelf 3D detectors completely fail.

\subsection{Viewpoint Distillation Module}

After establishing 2D-3D semantic and instance consistencies, the refined candidate set $\mathcal{O}'$ provides robust geometric priors. However, directly projecting these proposals onto all available video frames introduces severe spatial redundancy, overwhelming the token limits of downstream VLMs. To distill the most informative visual evidence, the Viewpoint Distillation module explicitly eliminates viewpoint redundancy through 3D spatial clustering of camera directions. Subsequently, by explicitly pairing the extracted optimal RGB frames with Bird's Eye View (BEV) maps, we construct concise visual prompt sets, effectively formulating the final target disambiguation as a multiple-choice reasoning task.

For each valid proposal $k \in \mathcal{O}'$, we first identify the subset of multi-view frames $\mathcal{F}_k \subset \mathcal{I}$ where the instance is visible. To ensure high-quality visual observations while minimizing computational overhead, we prioritize frames where the target is most prominently displayed. Specifically, we rank the visible frames based on the projected 2D bounding box area of the target, and select the top-$K_v$ frames to form a refined, high-quality observation subset $\tilde{\mathcal{F}}_k$.
% \begin{equation}
%     \tilde{\mathcal{F}}_k = \mathop{\text{Top-}K_v}_{f \in \mathcal{F}_k} \left( \text{Area}(f, B_k) \right),
% \end{equation}
% where $\text{Area}(f, B_k)$ computes the projected 2D bounding box area of the $k$-th target in frame $f$.
Subsequently, instead of exhaustively processing all visible frames, we strictly focus on this refined subset $\tilde{\mathcal{F}}_k$. For each frame $f \in \tilde{\mathcal{F}}_k$, we extract its associated camera pose and compute the normalized viewing direction vector $\mathbf{d}_f \in \mathbb{R}^3$ pointing towards the target instance. To group redundant, adjacent viewpoints, we measure the angular distance between any two camera vectors $\mathbf{d}_i$ and $\mathbf{d}_j$:
\begin{equation}
    D_{\text{angle}}(\mathbf{d}_i, \mathbf{d}_j) = \arccos \left( \frac{\mathbf{d}_i \cdot \mathbf{d}_j}{\| \mathbf{d}_i \| \| \mathbf{d}_j \|} \right).
\end{equation}
Based on this angular metric, we perform spatial clustering to partition the selected top-$K_v$ viewpoints of proposal $k$ into $C_k$ distinct clusters $\{ \Omega_{k,1}, \dots, \Omega_{k,C_k} \}$. A cluster $\Omega_{k,c}$ is formed such that the internal angular distance relative to its geometric center $\mathbf{d}_{\text{center}}^{(k,c)}$ remains below a predefined threshold $\epsilon$.
% \begin{equation}
%     \Omega_{k,c} = \left\{ f \in \tilde{\mathcal{F}}_k \mid D_{\text{angle}}(\mathbf{d}_f, \mathbf{d}_{\text{center}}^{(k,c)}) < \epsilon \right\}.
% \end{equation}
Finally, to represent each cluster $\Omega_{k,c}$ with the most informative single view, we straightforwardly select the optimal representative frame $f^*_{k,c}$ that maximizes the target's visible pixel area within that specific viewing direction.
% \begin{equation}
%     f^*_{k,c} = \arg\max_{f \in \Omega_{k,c}} \text{Area}(f, B_k).
% \end{equation}

To explicitly bridge the spatial gap and enforce strict 2D-3D viewpoint consistency, we introduce a Bird's Eye View (BEV) spatial mapping mechanism. For each selected video frame $f^*_{k,c}$ of proposal $k$, we render a corresponding local BEV map $I_{\text{BEV}}^{(k, c)}$. To provide the VLM with holistic geometric awareness and unambiguous instance referentiality, we explicitly overlay numeric ID markers on the target objects across the images. Simultaneously, we plot directional vector arrows on the BEV map, precisely indicating the camera's observation pose and viewing angle relative to the 3D scene. 

Subsequently, we spatially concatenate the RGB video frame with its corresponding BEV map to form a 2D-3D viewpoint visual prompt pair $p_{k,c}$:
\begin{equation}
    p_{k,c} = \left[ I_{f^*_{k,c}} \parallel I_{\text{BEV}}^{(k, c)} \right],
\end{equation}
where $\parallel$ denotes the side-by-side image concatenation operation. For a specific proposal $k$, all its generated visual prompt pairs across different viewpoints are concatenated together to construct a comprehensive viewpoint visual prompt set $\mathcal{V}_k$.
% \begin{equation}
%     \mathcal{V}_k = \bigcup_{c=1}^{C_k} p_{k,c}.
% \end{equation}

Ultimately, instead of conducting computationally expensive exhaustive spatial searches, we formulate the final target disambiguation as a multiple-choice reasoning task. We feed the viewpoint visual prompt sets from multiple candidate proposals $\mathcal{V} = \{ \mathcal{V}_k \}_{k=1}^{|\mathcal{O}'|}$, along with the original language query $\mathcal{Q}$, into the VLM. The VLM performs iterative cross-view and cross-proposal spatial reasoning to select the best-matching candidate proposal. Consequently, the 3D bounding box corresponding to this optimal proposal is extracted as the definitive grounded result $\mathcal{B}^*$.

\begin{table*}[htbp]
    \caption{Comparison with the state-of-the-art methods on ScanRefer in terms of different scene types.}
    \vspace{-0.3cm}
    \centering
    \renewcommand{\arraystretch}{1.2} 
    \resizebox{1.0\linewidth}{!}{
        \begin{tabular}{l|c|c|cc|cc|cc}
            \toprule
            \multirow{2}{*}{\textbf{Method}} & 
            \multirow{2}{*}{\textbf{Source}} & 
            \multirow{2}{*}{\textbf{Modality}} & 
            \multicolumn{2}{c|}{\textbf{Unique}} & 
            \multicolumn{2}{c|}{\textbf{Multiple}} & 
            \multicolumn{2}{c}{\textbf{Overall}} \\ 
            \cline{4-9}
             &  &  & 
            \textbf{Acc$@$0.25} & \textbf{Acc$@$0.5} & 
            \textbf{Acc$@$0.25} & \textbf{Acc$@$0.5} & 
            \textbf{Acc$@$0.25} & \textbf{Acc$@$0.5} \\ 
            \hline
            \multicolumn{9}{c}{\textbf{Fully-Supervised Methods}} \\ 
            \hline
            ScanRefer~\cite{chen2020scanrefer} & ECCV20 & 3D & 67.64 & 46.19 & 32.06 & 21.26 & 38.97 & 26.10 \\
            TGNN~\cite{huang2021text} & AAAI21 & 3D & 68.61 & 56.80 & 29.84 & 23.18 & 37.37 & 29.70 \\
            SAT~\cite{yang2021sat} & ICCV21 & 3D+2D & 73.21 & 50.83 & 37.64 & 25.16 & 44.54 & 30.14 \\
            MVT~\cite{huang2022multi} & CVPR22 & 3D+2D & 77.67 & 66.45 & 31.92 & 25.26 & 40.80 & 33.26 \\
            3D-SPS~\cite{luo20223d} & CVPR22 & 3D+2D & 84.12 & 66.72 & 40.32 & 29.82 & 48.82 & 36.98 \\
            EDA~\cite{wu2023eda} & CVPR23 & 3D & 85.76 & 68.57 & 49.13 & 37.64 & 54.59 & 42.26 \\
            3DVLP~\cite{zhang2024vision} & CVPR23 & 3D+2D & 84.23 & 64.61 & 43.51 & 33.41 & 51.41 & 39.46 \\
            MCLN~\cite{qian2024multi} & ECCV24 & 3D & 86.89 & 72.73 & 51.96 & 40.76 & 57.17 & 45.53 \\ 
            \hline
            \multicolumn{9}{c}{\textbf{Zero-Shot Methods}} \\ 
            \hline
            LERF~\cite{kerr2023lerf} & ICCV23 & 3D+2D & - & - & - & - & 4.8 & 0.9 \\
            OpenScene~\cite{peng2023openscene} & CVPR23 & 3D+2D & 20.1 & 13.1 & 11.1 & 4.4 & 13.2 & 6.5 \\
            LLM-Grounder~\cite{yang2024llm} & ICRA24 & 3D & - & - & - & - & 17.1 & 5.3 \\
            ZS3DVG~\cite{yuan2024visual} & CVPR24 & 3D+2D & 63.8 & 58.4 & 27.7 & 24.6 & 36.4 & 32.7 \\
            VLM-Grounder~\cite{xu2024vlm} & CoRL24 & 2D & 66.0 & 29.8 & 48.3 & 33.5 & 51.6 & 32.8 \\
            SeeGround~\cite{li2025seeground} & CVPR25 & 3D+2D & 75.7 & 68.9 & 34.0 & 30.0 & 44.1 & 39.4 \\
            SeqVLM~\cite{lin2025seqvlm} & ACMMM25 & 3D+2D & 77.3 & 72.7 & 47.8 & 41.3 & 55.6 & 49.6 \\ 
            % SeqVLM~\cite{} & ACMMM25 & 3D+2D & 77.3 & 72.7 & 47.8 & 41.3 & 55.6 & 49.6 \\ 
            \textbf{MCM-VG (ours)} & - & 3D+2D & \textbf{81.8} & \textbf{74.2} & \textbf{54.9} & \textbf{46.2} & \textbf{62.0}& \textbf{53.6}\\ 
            \bottomrule
        \end{tabular}
    }
    \vspace{-0.1cm}
    \label{tab:scanrefer_result}
\end{table*}

\begin{table}[htbp]
    \centering
    \renewcommand{\arraystretch}{1.2} % 行距
    \caption{Comparison with the state-of-the-art methods on Nr3D in terms of different evaluation metrics.}
    \vspace{-0.2cm}
    \resizebox{1.0\linewidth}{!}{
        \begin{tabular}{l|cc|cc|c}
            \toprule
            \textbf{Method} & \textbf{Easy} & \textbf{Hard} & \textbf{Dep.} & \textbf{Indep.} & \textbf{Overall} \\ 
            \hline
            \multicolumn{6}{c}{\textbf{Fully-Supervised Methods}} \\ 
            \hline
            ReferIt3D~\cite{achlioptas2020referit3d} & 43.6 & 37.9 & 32.5 & 37.1 & 35.6  \\
            TGNN~\cite{huang2021text} & 44.2 & 30.6 & 35.8 & 38.0 & 37.3 \\
            InstanceRefer~\cite{yuan2021instancerefer} & 46.0 & 31.8 & 34.5 & 41.9 & 38.8  \\
            3DVG-Trans~\cite{zhao20213dvg} & 48.5 & 34.8 & 34.8 & 43.7 & 40.8  \\
            SAT~\cite{yang2021sat} & 56.3 & 42.4 & 46.9 & 50.4 & 49.2  \\
            EDA~\cite{wu2023eda} & 58.2 & 46.1 & 50.2 & 53.1 & 52.1 \\
            MCLN~\cite{qian2024multi} & - & - & - & - & 59.8  \\ 
            \hline
            \multicolumn{6}{c}{\textbf{Zero-Shot Methods}} \\ 
            \hline
            ZS3DVG~\cite{yuan2024visual} & 46.5 & 31.7 & 36.8 & 40.0 & 39.0 \\ 
            SeeGround~\cite{li2025seeground} & 54.5 & 38.3 & 42.3 & 48.2 & 46.1 \\ 
            VLM-Grounder~\cite{xu2024vlm} & 55.2 & 39.5 & 45.8 & 49.4 & 48.0 \\ 
            SeqVLM~\cite{lin2025seqvlm} & 58.1 & 47.4 & 51.0 & 54.5 & 53.2 \\
            \textbf{MCM-VG (ours)} & \textbf{59.6} & \textbf{54.4} & \textbf{54.2} & \textbf{59.1} & \textbf{57.2} \\
            \bottomrule
        \end{tabular}
    }
    \vspace{-0.3cm}
    \label{tab:nr3d_result}
\end{table}

\section{Experiments}
\subsection{Experimental Settings}
\textbf{Datasets.} We conduct experiments on two standard benchmarks for 3D visual grounding: ScanRefer \cite{chen2020scanrefer} and Nr3D \cite{achlioptas2020referit3d}. Derived from ScanNet, ScanRefer provides 51,583 referring expressions for 11,046 objects in 800 indoor scenes. This benchmark mainly focuses on fine-grained grounding, requiring models to associate language descriptions with 3D targets by exploiting object attributes as well as spatial relations. Following the official setup, its test set is further divided into Unique and Multiple cases. The Unique subset contains objects that can be identified directly from their own properties, whereas the Multiple subset involves scenes with several instances from the same semantic category, making relational reasoning essential for disambiguation.
Nr3D places greater emphasis on natural, human-generated utterances and the diversity of viewpoint-aware descriptions in real indoor environments. It contains 41,503 expressions referring to 7,189 objects across 1,448 scenes. To provide a more detailed analysis, the benchmark evaluates performance under several splits, including Easy and Hard, which reflect the difficulty of contextual inference, as well as View-Dependent and View-Independent, which examine whether successful grounding relies on the observer’s perspective.

\noindent\textbf{Implementation Details.}
% Experiments were conducted on a RTX-3090 GPU using GPT-5 as the LLM assisted with CLIP-ViT-Base-Patch16 \cite{} text, vision encoder for Semantic Alignment and Doubao-Seed-1.6-vision \cite{} as the VLM for multimodal reasoning. 
All experiments were conducted on an RTX-3090 GPU. We use GPT-5 \cite{singh2025openai} as the LLM for parsing the free-form query, and Doubao-Seed-1.6-vision \cite{bytedance_doubao16} as the VLM for multimodal reasoning. 
We employ Mask3D \cite{schult2023mask3d} and SAM3 \cite{carion2025sam} as the 3D and 2D semantic segmentation models, respectively.
For hyperparameters, $\gamma$, $\epsilon$, and $k_v$ are set to 0.07, 30, and 5, respectively.
We divide each scene video into 20-frame segments for viewpoint sampling, and incorporate Viewpoint Distillation, ultimately retaining a few optimal viewpoints and their corresponding BEV maps.
The VLM processes candidates through an iterative reasoning mechanism with a batch size limit of four. 
Due to the high computational cost of VLM-based models, we follow the standardized protocol of VLM-Grounder\cite{xu2024vlm} and SeqVLM\cite{lin2025seqvlm} for fair evaluation and reproducibility, testing on 250 validation samples per benchmark dataset, matching previous evaluation conditions.

\begin{table*}[t]
    \caption{Ablation studies of different components and strategies on ScanRefer. SA, IR, and VD denote the Semantic Alignment Module, the Instance Rectification Module, and the Viewpoint Distillation Module, respectively.}
    \centering

    %==================== first row ====================%
    \begin{minipage}[t]{0.48\textwidth}
        \centering
        \setlength{\tabcolsep}{8pt}
        \renewcommand{\arraystretch}{1.1}
        \subcaption{Ablation on different components in our framework.}
        % \vspace{0.3em}

        \begin{threeparttable}
            \begin{tabular}{cccc|cc}
                \toprule
                \textbf{Baseline} & \textbf{SA} & \textbf{IR} & \textbf{VD} & \textbf{Acc$@$0.25} & \textbf{Acc$@$0.5} \\
                \midrule
                \checkmark &  &  &  & 54.8 & 48.4 \\
                \checkmark &  & \checkmark & \checkmark & 58.0 & 50.8 \\
                \checkmark & \checkmark &  & \checkmark & 57.2 & 50.4 \\
                \checkmark & \checkmark & \checkmark &  & 57.6 & 50.0 \\
                \rowcolor{gray!20}
                \checkmark & \textbf{\checkmark} & \checkmark & \checkmark & \textbf{62.0} & \textbf{53.6} \\
                \bottomrule
            \end{tabular}
        \end{threeparttable}
        \label{tab:ablation-components}
    \end{minipage}
    \hfill
    \begin{minipage}[t]{0.48\textwidth}
        \centering
        \setlength{\tabcolsep}{8pt}
        \renewcommand{\arraystretch}{1.4}
        \subcaption{Ablation on different strategies in SA.}
        % \vspace{0.3em}

        \begin{tabular}{l|cc}
            \toprule
            \textbf{Different Strategies} & \textbf{Acc$@$0.25} & \textbf{Acc$@$0.5} \\
            \midrule
            Top-5 Category Proposals & 60.8 & 53.2 \\
            \rowcolor{gray!20}
            Top-10 Category Proposals & \textbf{62.0} & \textbf{53.6} \\
            Top-15 Category Proposals & 61.6 & 52.8 \\
            % w.o. 2D Segmentation Model & 58.0 & 50.8 \\
            \bottomrule
        \end{tabular}
        \label{tab:ablation-sa}
    \end{minipage}

    \vspace{0.8em}

    %==================== second row ====================%
    \begin{minipage}[t]{0.48\textwidth}
        \centering
        \setlength{\tabcolsep}{7pt}
        \renewcommand{\arraystretch}{1.4}
        \subcaption{Ablation on different strategies in IR.}
        % \vspace{0.3em}

        \begin{tabular}{l|cc}
            \toprule
            \textbf{Different Strategies} & \textbf{Acc$@$0.25} & \textbf{Acc$@$0.5} \\
            \midrule
            Point Prompts (1 pos.) & 58.2 & 49.2 \\
            Point Prompts (1 pos. \& 1 neg.) & 59.4 & 51.6 \\
            \rowcolor{gray!20}
            Point Prompts (2 pos. \& 1 neg.) & \textbf{62.0} & \textbf{53.6} \\
            % w.o. 2D Segmentation Model & 57.2 & 50.4 \\
            \bottomrule
        \end{tabular}
        \label{tab:ablation-ir}
    \end{minipage}
    \hfill
    \begin{minipage}[t]{0.48\textwidth}
        \centering
        \setlength{\tabcolsep}{10pt}
        \renewcommand{\arraystretch}{1.4}
        \subcaption{Ablation on different strategies in VD.}
        % \vspace{0.3em}

        \begin{tabular}{l|cc}
            \toprule
            \textbf{Different Strategies} & \textbf{Acc$@$0.25} & \textbf{Acc$@$0.5} \\
            \midrule
            w.o. Spatial Clustering & 58.4 & 50.0 \\
            w.o. BEV Spatial Mapping & 59.6 & 52.4 \\
            \rowcolor{gray!20}
            Full Model & \textbf{62.0} & \textbf{53.6} \\
            \bottomrule
        \end{tabular}
        \label{tab:ablation-vd}
    \end{minipage}
\end{table*}

\subsection{Comparison with State-of-the-Art Methods}
Table~\ref{tab:scanrefer_result} and \ref{tab:nr3d_result} present the quantitative comparison between our proposed MCM-VG and existing state-of-the-art methods on the ScanRefer and Nr3D benchmarks, respectively. 
% The results are extensively evaluated under ``Unique" (single target of its class) and ``Multiple" (multiple targets of the same class) scene categories, using Acc@0.25 and Acc@0.5 metrics.
As shown in the tables, MCM-VG sets a new state-of-the-art across all metrics in the zero-shot setting. 
For ScanRefer, our method achieves Overall Acc@0.25 and Acc@0.5 scores of 62.0\% and 53.6\%, outperforming the previous best zero-shot method, SeqVLM~\cite{lin2025seqvlm}, by substantial absolute margins of 6.4\% and 4.0\%, respectively. More impressively, in the notoriously challenging "Multiple" subset, which strictly requires fine-grained spatial disambiguation, MCM-VG yields 54.9\% (Acc@0.25) and 46.2\% (Acc@0.5), surpassing SeqVLM by an impressive 7.1\% and 4.9\%. 
Similarly, this robust generalization extends to the Nr3D benchmark. MCM-VG achieves Overall accuracy of 57.2\%, surpassing the strongest zero-shot competitor, SeqVLM, by a solid margin of 4.0\%. Furthermore, in the highly demanding "Hard" subset, which involves complex relational reasoning among multiple distractor objects, our method secures a 54.4\% accuracy, outperforming the baseline by 7.0\%. 
This significant leap explicitly validates the effectiveness of our framework, proving that providing VLMs with meticulously curated 2D-3D spatial prompts is superior to performing reasoning over redundant multi-view sequences. Remarkably, despite being training-free, MCM-VG surpasses some fully-supervised methods without relying on any task-specific annotations. This performance demonstrates that explicitly enforcing 2D-3D consistencies across semantics, instances, and viewpoints effectively unleashes the potential of foundation models, successfully bridging the performance gap between zero-shot and fully-supervised paradigms.

\begin{figure*}
    \centering
    \includegraphics[width=0.97\linewidth]{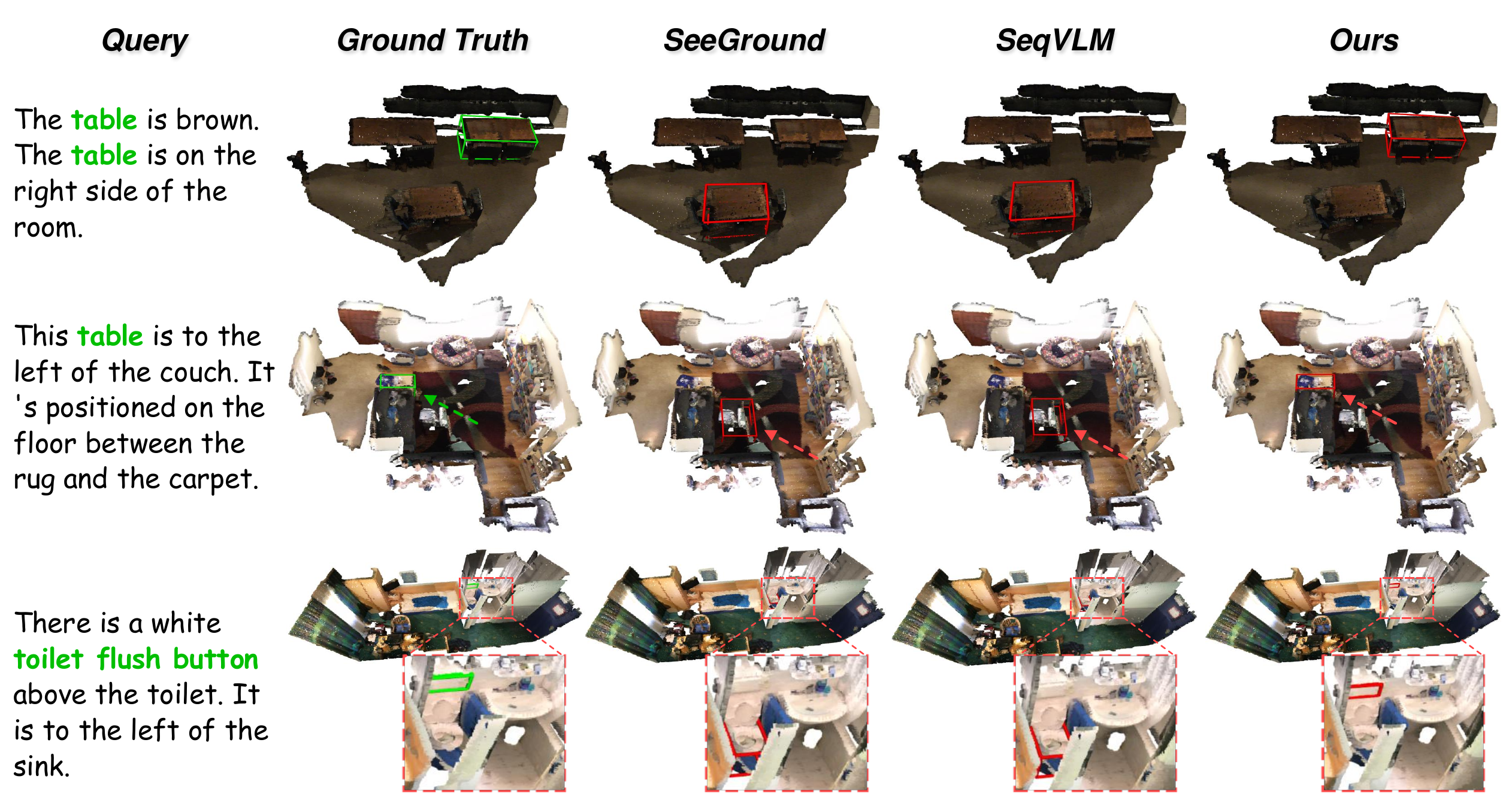}
    \vspace{-0.3cm}

    \caption{Visualization results of 3D visual grounding on the ScanRefer\cite{chen2020scanrefer} dataset.}
    
    \label{fig:visualize}
    \vspace{-0.1cm}
\end{figure*}

As illustrated in Fig.~\ref{fig:visualize}, we present a qualitative comparison between our MCM-VG and recent state-of-the-art zero-shot methods, namely SeeGround~\cite{li2025seeground} and SeqVLM~\cite{lin2025seqvlm}. Visually, our approach demonstrates markedly superior grounding performance. Notably, in highly cluttered and complex scenes, MCM-VG successfully localizes small or inconspicuous objects (Row 2-3). In such challenging scenarios, initial 3D proposals frequently suffer from erroneous category predictions or completely fail to generate bounding boxes for these diminutive targets, misleading the baseline methods. However, our proposed Semantic Alignment and Instance Rectification modules effectively compensate for these inherent deficiencies. By explicitly resolving semantic ambiguities and reconstructing missing geometries via reliable 2D visual priors, our framework accurately grounds the challenging targets where others fail, further validating its robustness and overall superiority.

\begin{figure}[t]
    \centering

    \begin{subfigure}[t]{0.32\linewidth}
        \centering
        \includegraphics[width=\linewidth]{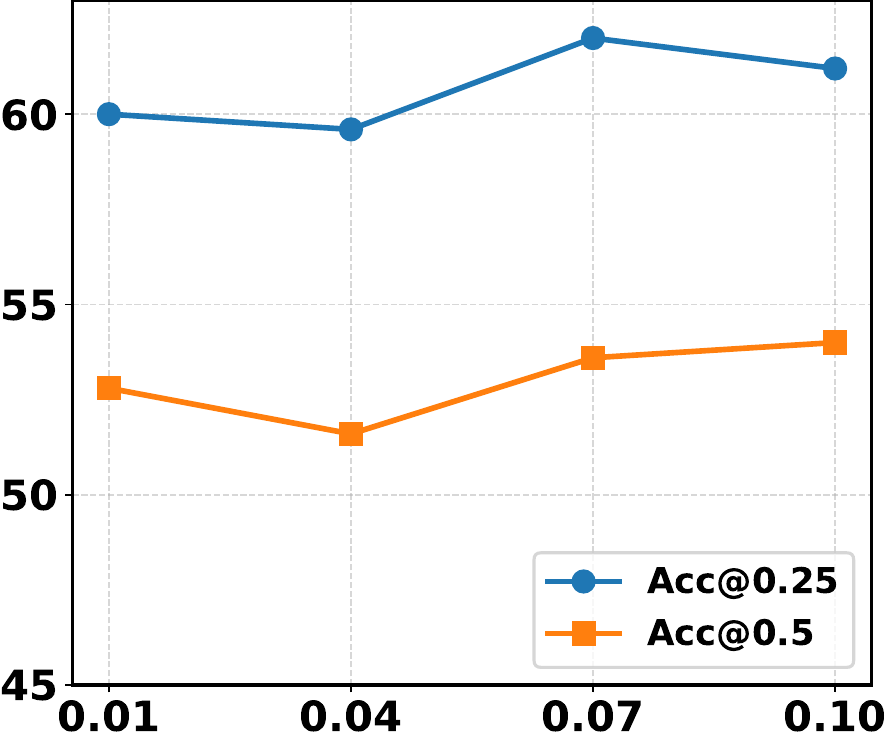}
        \caption{Effect of $\gamma$}
        \label{fig:ablation_gamma}
    \end{subfigure}\hfill
    \begin{subfigure}[t]{0.32\linewidth}
        \centering
        \includegraphics[width=\linewidth]{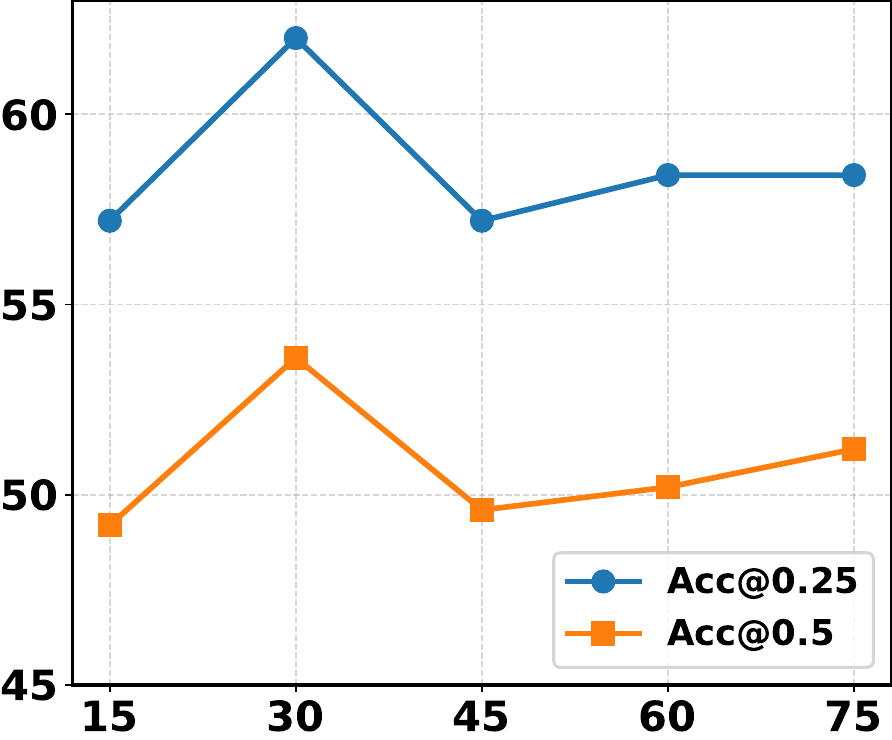}
        \caption{Effect of $\epsilon$}
        \label{fig:ablation_epsilon}
    \end{subfigure}\hfill
    \begin{subfigure}[t]{0.32\linewidth}
        \centering
        \includegraphics[width=\linewidth]{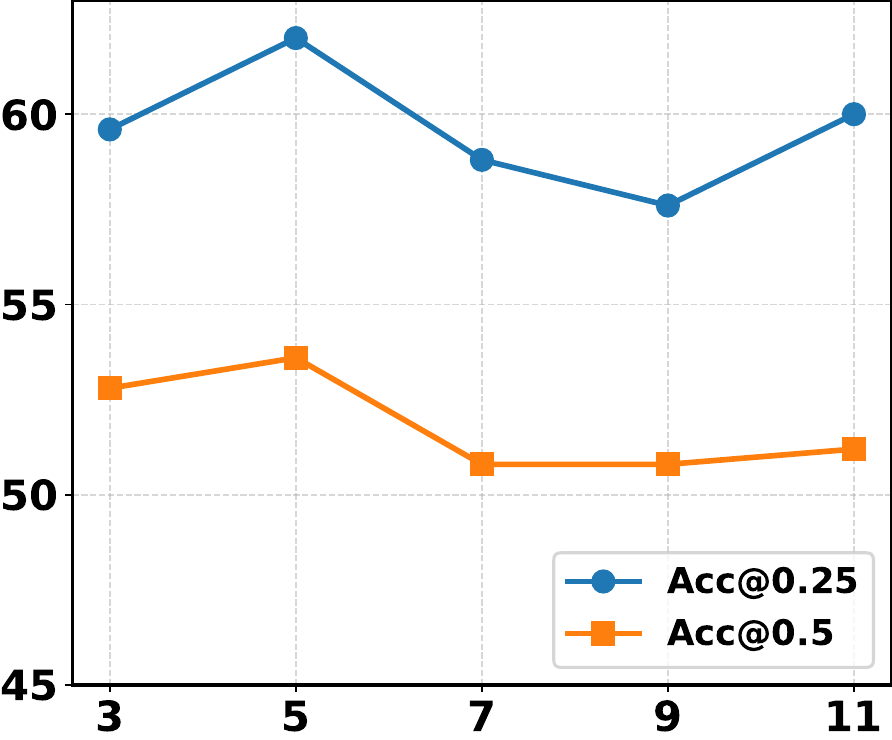}
        \caption{Effect of $K_v$}
        \label{fig:ablation_kv}
    \end{subfigure}

    \vspace{-0.2cm}
    \caption{Ablation studies of different hyperparameters.}
    \vspace{-0.3cm}
    \label{fig:ablation_all}
\end{figure}

\subsection{Ablation Studies}
\subsubsection{Components.} To validate the individual contributions of our proposed components, we conduct an extensive ablation study on the ScanRefer benchmark, as summarized in Table~\ref{tab:ablation-components}. The bare baseline bypasses our consistency mappings, directly feeding unrefined 3D proposals and exhaustive multi-view projections into the VLM, which yields a sub-optimal 54.8\% at Acc@0.25. Removing the Instance Rectification (IR) module results in a 4.8\% drop in Acc@0.25, validating our core hypothesis that explicitly leveraging 2D visual priors to rectify geometrically inaccurate 3D boundaries is indispensable for precise spatial localization. Similarly, omitting the Viewpoint Distillation (VD) module leads to a significant decrease of 4.4\%, demonstrating that without spatial clustering and Bird's Eye View (BEV) spatial prompting, the VLM is easily overwhelmed by the severe spatial redundancy of multi-view frames. Furthermore, the exclusion of the Semantic Alignment (SA) module yields a 4.0\% drop, clearly indicating that our coarse-to-fine 2D-3D matching mechanism is crucial for filtering out the open-vocabulary category ambiguities inherent in off-the-shelf 3D segmentation models. Ultimately, integrating all three modules collaboratively establishes the full MCM-VG pipeline, achieving the best performance of 62.0\% on Acc@0.25 and 53.6\% on Acc@0.5. This massive gain of 7.2\% over the baseline robustly proves that semantic, instance, and viewpoint consistencies are mutually reinforcing, establishing a highly accurate zero-shot 3D visual grounding paradigm.

% \begin{figure}
%     \centering
%     \includegraphics[width=0.98\linewidth]{figs/ablation_hyper_vd.pdf}
%     % \vspace{-0.5em}
%     \caption{Ablation on hyper-parameters of Viewpoint Distillation Module.}
%     \label{fig:intro}
%     \vspace{-1.5em}
% \end{figure}

\subsubsection{Semantic Alignment Module.}
% \textit{\noindent\textbf{Strategies:}} To validate the internal mechanisms of the SA module, we ablate its core strategies in Table~\ref{tab:ablation-sa}. Removing the 2D segmentation model incurs a severe 4.0\% drop in Acc@0.25 (from 62.0\% to 58.0\%). This clearly confirms that relying solely on coarse vision-language features (e.g., CLIP) is insufficient; explicit fine-grained 2D-3D semantic alignment is crucial for rigorously resolving category ambiguities. Furthermore, we investigate the number of LLM-generated category proposals. Retaining the Top-10 proposals yields the optimal performance. A smaller pool (Top-5) overly restricts the semantic search space, potentially discarding the correct target category, whereas a larger pool (Top-15) introduces excessive semantic noise, which inevitably distracts the coarse-to-fine matching process.
\textit{\noindent\textbf{Strategies:}} To validate the internal mechanisms of the Semantic Alignment (SA) module, we investigate the impact of the number of LLM-generated category proposals in Table~\ref{tab:ablation-sa}. Retaining the Top-10 proposals yields the optimal performance, achieving a peak Acc@0.25 of 62.0\%. A smaller candidate pool (Top-5) overly restricts the semantic search space, potentially discarding the correct target category during the initial coarse filtering phase. Conversely, expanding the pool to Top-15 introduces excessive semantic noise and ambiguous distractors, which inevitably interferes with the subsequent fine-grained 2D-3D matching process.
\textit{\noindent\textbf{Hyperparameters:}} We further evaluate the matching score threshold $\gamma$ in Fig.~\ref{fig:ablation_gamma}, which determines whether to accept the 2D-refined category. The optimal overall performance is achieved at $\gamma=0.07$, yielding a peak Acc@0.25 of 62.0\%. Setting a lower threshold yields sub-optimal results, as a loose criterion blindly accepts unreliable 2D-3D geometric matches, injecting false categorical updates into the 3D proposals. Conversely, an overly strict threshold conservatively rejects valid semantic alignments. This prevents the module from rectifying inherently noisy 3D categories. Nevertheless, the overall grounding accuracy remains robust across a wide range of $\gamma$ values.

\subsubsection{Instance Rectification Module.}
% We ablate the strategies of the Instance Rectification (IR) module in Table~\ref{tab:ablation-ir}. First, discarding the 2D segmentation model (4th row) leads to a 4.0\% drop in Acc@0.25. This confirms that explicitly refining boundaries via 2D segmentation models (e.g., SAM3) is indispensable for rectifying flawed 3D geometries. Furthermore, we investigate the impact of the VLM-generated spatial prompts used to guide SAM3. Providing only a single positive point yields a sub-optimal 58.2\%, as it often introduces segmentation ambiguity for complex objects. Introducing a negative point to explicitly exclude background interference improves the accuracy to 59.4\%. Ultimately, leveraging a richer prompt combination (2 positive \& 1 negative points) yields the peak performance of 62.0\%. This demonstrates that comprehensive, VLM-guided multi-point cues are critical for IR module.
To validate the internal mechanisms of the Instance Rectification (IR) module, we investigate the impact of different VLM-generated spatial prompts used to guide the 2D segmentation model, as shown in Table~\ref{tab:ablation-ir}. Providing only a single positive point yields a sub-optimal Acc@0.25 of 58.2\%, as such sparse guidance often introduces segmentation ambiguity when dealing with complex or partially occluded objects. Introducing a negative point to explicitly exclude background interference improves the accuracy to 59.4\%. Ultimately, leveraging a richer prompt combination (2 positive \& 1 negative points) achieves the peak performance of 62.0\%. This demonstrates that comprehensive, VLM-guided multi-point cues are critical for IR module.

% \begin{figure*}[t]
%     \centering

%     \begin{subfigure}[t]{0.28\textwidth}
%         \centering
%         \includegraphics[width=\linewidth]{figs/ablation_epsilon.pdf}
%         \caption{Effect of component A}
%         \label{fig:ablation_a}
%     \end{subfigure}\hspace{0.03\textwidth}
%     \begin{subfigure}[t]{0.28\textwidth}
%         \centering
%         \includegraphics[width=\linewidth]{figs/ablation_epsilon.pdf}
%         \caption{Effect of $\epsilon$}
%         \label{fig:ablation_epsilon}
%     \end{subfigure}\hspace{0.03\textwidth}
%     \begin{subfigure}[t]{0.28\textwidth}
%         \centering
%         \includegraphics[width=\linewidth]{figs/ablation_kv.pdf}
%         \caption{Effect of $K_v$}
%         \label{fig:ablation_kv}
%     \end{subfigure}

%     \caption{Ablation studies of different hyperparameters.}
%     \label{fig:ablation_all}
% \end{figure*}

\begin{table}[t]
    \caption{Ablation on different VLM models on ScanRefer.}
    \vspace{-0.25cm}
    \centering
    \setlength{\tabcolsep}{8.0pt} 
    \renewcommand{\arraystretch}{1.3}
    \begin{tabular}{l | l c c} 
    \toprule
    \textbf{Method} & \textbf{VLM Model} & \textbf{Acc$@$0.25} & \textbf{Acc$@$0.5} \\
    \midrule
    SeqVLM~\cite{lin2025seqvlm} & Doubao-1.6 & 54.8 & 48.4 \\
    \midrule
    Ours & Qwen2-VL-72B & 50.4 & 43.2 \\
    ~    & GPT-4o       & 55.6 & 48.8 \\
    \rowcolor{gray!20}
    ~    & \textbf{Doubao-1.6} & \textbf{62.0} & \textbf{53.6} \\
    \bottomrule
    \end{tabular}
    \vspace{-0.2cm}
    \label{tab:ablation_vlm}
\end{table}

\subsubsection{Viewpoint Distillation Module}
\textit{\noindent\textbf{Strategies:}} To further dissect the efficacy of our Viewpoint Distillation module, we ablate its two core strategies: Spatial Clustering and Bird's Eye View (BEV) Spatial Mapping. As reported in Table~\ref{tab:ablation-vd}, both components are indispensable. Omitting the Spatial Clustering mechanism incurs the most severe performance penalty, dropping Acc@0.25 by a substantial 3.6\% (from 62.0\% to 58.4\%). Without angular clustering to eliminate redundant viewpoints, the VLM is fed a chaotic sequence of highly overlapping frames, which inevitably distracts its attention and overwhelms its limited reasoning capacity. Furthermore, removing the BEV Spatial Mapping degrades Acc@0.25 by 2.4\% (from 62.0\% to 59.6\%). This highlights that relying solely on isolated 2D RGB frames deprives the VLM of holistic top-down geometric awareness, making it exceedingly difficult to perform accurate cross-view spatial reasoning without explicit camera directional cues. Ultimately, the full model synergizes clustering-based redundancy elimination with BEV-guided spatial prompting, achieving the peak accuracy.
\textit{\noindent\textbf{Hyperparameters:}} As illustrated in Fig.~\ref{fig:ablation_epsilon} and \ref{fig:ablation_kv}, we investigate the impact of the multi-view frame number $K_v$ and the angular clustering threshold $\epsilon$. Our framework maintains highly competitive performance across a broad range of parameter values. For $K_v$, the performance peaks at $K_v=5$ (62.0\% on Acc@0.25). Fewer frames ($K_v=3$) slightly degrade accuracy due to insufficient visual diversity, while introducing excessive frames ($K_v \geq 7$) inevitably includes occluded or marginal observations that distract the VLM. Similarly, the optimal angular threshold is observed at $\epsilon=30^\circ$. A stricter threshold ($\epsilon=15^\circ$) causes under-clustering and feeds redundant viewpoints to the VLM, whereas a looser threshold ($\epsilon \geq 45^\circ$) leads to over-clustering, forcibly merging distinct and valuable spatial evidence. Despite these trade-offs, the overall accuracy remains remarkably stable across broad settings, underscoring the robustness of our viewpoint distillation module.

\subsubsection{VLM Models.} 
% To evaluate the generalization of our framework, we ablate the choice of the Vision-Language Model (VLM) used for the final spatial reasoning stage. As shown in Table~\ref{tab:ablation_vlm}, our method maintains competitive performance across various state-of-the-art VLMs, demonstrating strong model-agnostic robustness. While the open-weight Qwen2-VL-72B \cite{wang2024qwen2} and the proprietary GPT-4o \cite{hurst2024gpt} achieve solid Acc@0.25 scores of 50.4\% and 55.6\% respectively, Doubao-1.6 yields the peak performance of 62.0\% on Acc@0.25 and 53.6\% on Acc@0.5, surpassing GPT-4o by a substantial absolute margin of 6.4\%. This significant leap indicates that Doubao-1.6 possesses superior fine-grained spatial comprehension and robust multiple-choice reasoning capabilities when interpreting our meticulously curated RGB and BEV visual prompt groups. Consequently, we adopt Doubao-1.6 as the default VLM engine in our framework.
We investigate the impact of different Vision-Language Model (VLM) used for final spatial reasoning. As shown in Table~\ref{tab:ablation_vlm}, our method maintains strong model-agnostic robustness. While Qwen2-VL-72B \cite{wang2024qwen2} and GPT-4o \cite{hurst2024gpt} achieve solid Acc@0.25 scores of 50.4\% and 55.6\%, Doubao-1.6 yields the peak performance of 62.0\%, indicating its superior fine-grained spatial comprehension and visual reasoning capabilities. Crucially, when equipped with the same Doubao-1.6 engine, our framework outperforms SeqVLM \cite{lin2025seqvlm} (54.8\%) by a massive 7.2\% absolute margin, which further demonstrates the effectiveness of our method.

\section{Conclusion}
% In this paper, we introduced MCM-VG, a novel and robust framework that tackles the fundamental bottlenecks of zero-shot 3D visual grounding: noisy open-vocabulary proposals and redundant multi-view reasoning. Rather than passively relying on flawed 3D segments, our approach explicitly enforces rigorous cross-modal alignment across semantic, instance, and viewpoint levels. Through the synergistic integration of the Semantic Alignment, Instance Rectification, and Viewpoint Distillation modules, MCM-VG effectively rectifies categorical and geometric inaccuracies. Furthermore, by distilling concise RGB-BEV visual prompt sets, it formulates complex 3D grounding as a multiple-choice reasoning task for VLMs. Extensive experiments on the ScanRefer and Nr3D benchmarks demonstrate the effectiveness of our framework.
In this paper, we introduce MCM-VG, a robust framework tackling the fundamental bottlenecks of zero-shot 3D visual grounding: noisy open-vocabulary proposals and redundant multi-view reasoning. Rather than passively relying on flawed 3D segments, our approach explicitly enforces rigorous 2D-3D consistency across semantic, instance, and viewpoint levels. Specifically, the Semantic Alignment and Instance Rectification modules synergistically rectify categorical and geometric inaccuracies. Furthermore, the Viewpoint Distillation module distills concise RGB-BEV visual prompt sets, formulating complex 3D grounding as a multiple-choice reasoning task for VLMs. Extensive experiments on the ScanRefer and Nr3D benchmarks demonstrate the effectiveness of our framework.
\bibliographystyle{ACM-Reference-Format}
\bibliography{sample-base}

\appendix

\section{Overview}
In this supplementary material, we provide comprehensive technical details, extended analyses, and additional qualitative results to complement the main manuscript. Specifically, \textbf{Section \ref{details}} formalizes the step-by-step execution flow of our MCM-VG framework via a detailed pseudo-code algorithm. \textbf{Section \ref{limitation}} discusses the inherent limitations of our current approach and outlines potential avenues for future research. \textbf{Section \ref{ir}} elaborates on the implementation details and ablation studies within the Instance Rectification Module. \textbf{Section \ref{vd}} presents a detailed ablation study on the Viewpoint Distillation module, demonstrating its effectiveness in pruning visual redundancy and enhancing final grounding accuracy. \textbf{Section \ref{vis}} showcases more qualitative visualization results across diverse and challenging scenarios. Finally, to ensure full transparency and reproducibility, \textbf{Section \ref{prompt}} comprehensively details the exact textual prompts employed throughout our modules.

\section{Detailed Pipeline of MCM-VG}
\label{details}
To provide a more comprehensive and step-by-step understanding of our proposed framework, we summarize the systematic execution flow of MCM-VG in Algorithm \ref{alg:mcm_vg}. The overall pipeline is structurally divided into three cohesive phases. Phase 1 establishes strict 2D-3D semantic consistency, explicitly verifying and filtering noisy 3D proposals via cross-modal alignment. Crucially, to handle extreme scenarios where off-the-shelf 3D detectors completely miss the target, Phase 2 functions as a robust fallback mechanism, reconstructing missing geometries directly from VLM-guided 2D segmentations. Finally, Phase 3 distills redundant multi-view observations into representative RGB-BEV prompt pairs via spatial clustering, elegantly formulating the final target disambiguation as an efficient multiple-choice reasoning task.

 % \vspace{-0.3cm}

\begin{algorithm}[htbp]
\small
\caption{Overall Pipeline of MCM-VG}
\label{alg:mcm_vg}
\SetAlgoLined
\KwIn{Point cloud $\mathcal{P}$, multi-view frames $\mathcal{I}$, language query $\mathcal{Q}$.}
\KwOut{Definitive grounded 3D bounding box $\mathcal{B}^*$.}

\tcp{\textbf{Phase 1: Semantic Alignment}}
$\mathcal{O} \leftarrow \Phi_{\text{3D}}(\mathcal{P})$ \tcp*{Initial 3D proposals}
Select top-3 categories $\mathcal{C}_{\text{top-3}}$ via LLM parsing and CLIP similarity\;
Initialize $\mathcal{O}' \leftarrow \emptyset$, $\mathcal{C}_{\text{valid}} \leftarrow \emptyset$\;
\For{proposal $k$ where $C_k \in \mathcal{C}_{\text{top-3}}$}{
    Compute spatial matching score $\eta_k$ via $\Psi_{\text{SEG}}$ and Eq.(5)\;
    \If{$\eta_k \geq \gamma$}{
        $\mathcal{C}_{\text{valid}} \leftarrow \mathcal{C}_{\text{valid}} \cup \{C_k\}$\;
        $\mathcal{O}' \leftarrow \mathcal{O}' \cup \{(M_k, B_k, T_{\text{target}})\}$\;
    }
}
Salvage unmatched proposals into $\mathcal{O}'$ if $C_k \in \mathcal{C}_{\text{valid}}$\;

\tcp{\textbf{Phase 2: Instance Rectification}}
\If{$\max \eta_k < \gamma$ \tcp*{Off-the-shelf 3D fails}}{
    $\mathcal{O}' \leftarrow \emptyset$\;
    \For{proposal $k \in \text{top-3 subsets}$}{
        $\hat{P}_{k,f} \leftarrow \text{VLM}(\mathcal{I}_k, \mathcal{Q}, \text{anchors})$\;
        $\hat{B}_{k,f} \leftarrow \Psi_{\text{SEG}}(I_f, \hat{P}_{k,f})$\;
        $M^*_k, B^*_k \leftarrow f_{\text{back-proj}}(\{\hat{B}_{k,f}\}, \mathcal{P})$\;
        $\mathcal{O}' \leftarrow \mathcal{O}' \cup \{(M^*_k, B^*_k, T_{\text{target}})\}$\;
    }
}

\tcp{\textbf{Phase 3: Viewpoint Distillation}}
Initialize visual prompt set $\mathcal{V} \leftarrow \emptyset$\;
\For{$k \in \mathcal{O}'$}{
    Select top-$K_v$ visible frames $\tilde{\mathcal{F}}_k$\;
    Cluster views into $C_k$ groups based on angular distance $D_{\text{angle}} < \epsilon$\;
    \For{each cluster $c \in \{1, \dots, C_k\}$}{
        Select optimal frame $f^*_{k,c}$ and render BEV map $I_{\text{BEV}}^{(k,c)}$\;
        Prompt pair $p_{k,c} \leftarrow \left[ I_{f^*_{k,c}} \parallel I_{\text{BEV}}^{(k, c)} \right]$\;
        Add $p_{k,c}$ to proposal visual set $\mathcal{V}_k$\;
    }
    $\mathcal{V} \leftarrow \mathcal{V} \cup \{\mathcal{V}_k\}$\;
}
$\mathcal{B}^* \leftarrow \text{VLM\_MultipleChoice}(\mathcal{V}, \mathcal{Q})$\;
\KwRet{$\mathcal{B}^*$}\;
\end{algorithm}

\begin{table*}[t]
    \centering
    \caption{More ablation studies on the Instance Rectification (IR) module. SEG denotes the 2D segmentation model.}
    % \vspace{-0.3cm}
    \setlength{\tabcolsep}{6pt}
    \renewcommand{\arraystretch}{1.4}
    \small

    \begin{subtable}[t]{0.33\textwidth}
        \centering
        \subcaption{Different VLMs.}
        \label{tab:ablation-vlms}
        \begin{tabular}{lcc}
            \toprule
            \textbf{VLM Model} & \textbf{Acc$@$0.25} & \textbf{Acc$@$0.5} \\
            \midrule
            GPT-5   & 59.2 & 52.0 \\
            Qwen3.5-plus  & 60.0 & 51.6 \\
            \rowcolor{gray!20}
            \textbf{Gemini-3-pro} & \textbf{62.0} & \textbf{53.6} \\
            \bottomrule
        \end{tabular}
    \end{subtable}
    % \hspace{0.005\textwidth}
    \begin{subtable}[t]{0.33\textwidth}
        \centering
        \subcaption{Different proposal generation strategies.}
        \label{tab:ablation-strategies}
        \begin{tabular}{cccc}
            \toprule
            \textbf{VLM} & \textbf{SEG} & \textbf{Acc$@$0.25} & \textbf{Acc$@$0.5} \\
            \midrule
            Boxes  & --     &     58.8     &      50.8    \\
            --     & Text   &     58.4     &      50.8    \\
            \rowcolor{gray!20}
            \textbf{Points} & \textbf{Points} & \textbf{62.0} & \textbf{53.6} \\
            \bottomrule
        \end{tabular}
    \end{subtable}
    % \hspace{0.005\textwidth}
    \begin{subtable}[t]{0.33\textwidth}
        \centering
        \subcaption{Different sources of proposals.}
        \label{tab:ablation-sources}
        \begin{tabular}{lcc}
            \toprule
            \textbf{Proposals} & \textbf{Acc$@$0.25} & \textbf{Acc$@$0.5} \\
            \midrule
            2D only & 20.6 & 8.9 \\
            3D only & 57.2 & 50.4 \\
            \rowcolor{gray!20}
            \textbf{Ours (3D+2D)} & \textbf{62.0} & \textbf{53.6} \\
            \bottomrule
        \end{tabular}
    \end{subtable}
\end{table*}

% \section{Limitations and Failure Cases}
\section{Limitations and Future Work}
\label{limitation}
\begin{figure}
    \centering
    \includegraphics[width=0.98\linewidth]{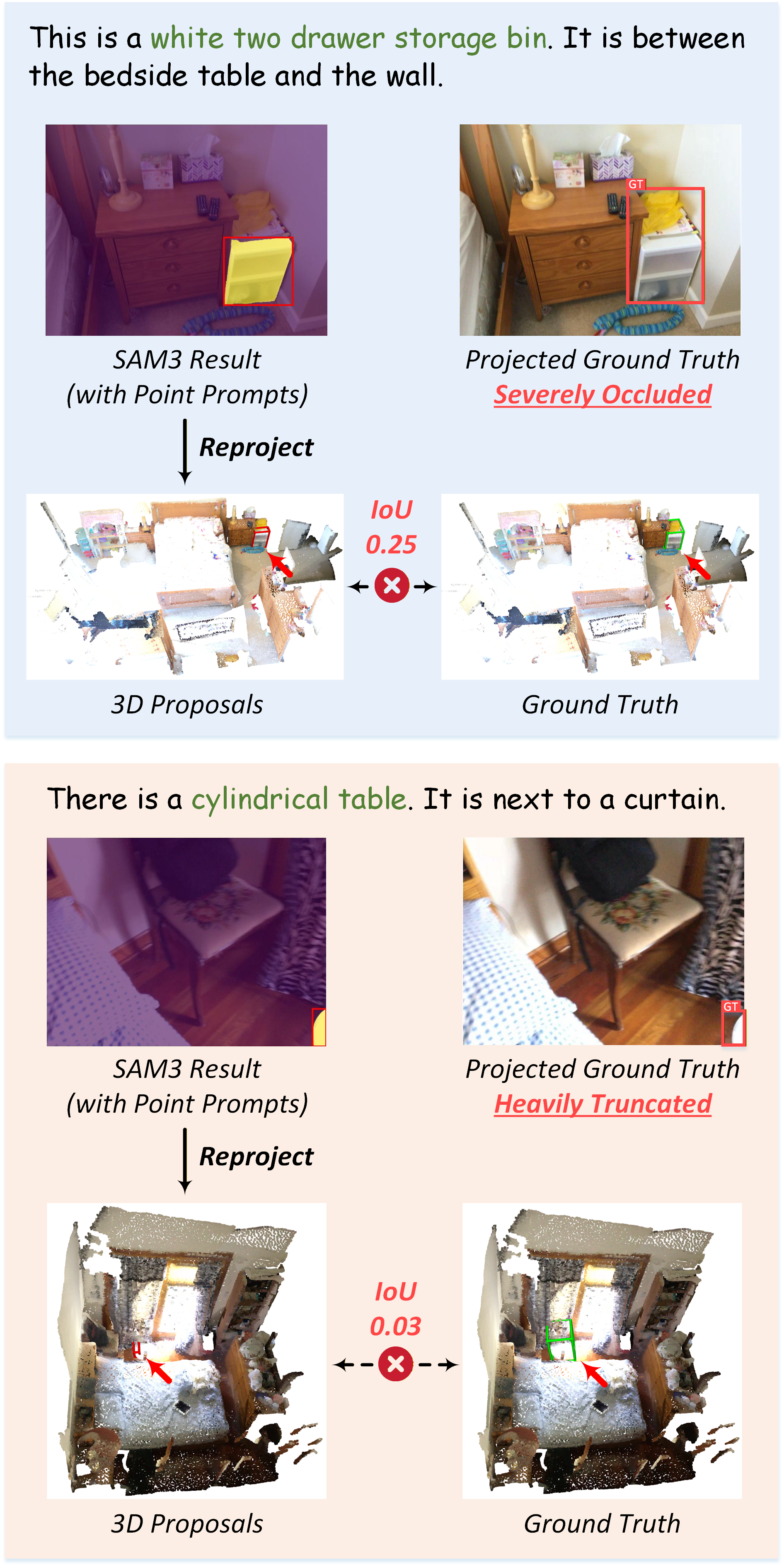}
    \vspace{-0.5em}
    \caption{Failure cases of our framework under severe occlusion and heavy view truncation.}
    \label{fig:failure}
    \vspace{-0.3cm}
\end{figure}

While MCM-VG demonstrates state-of-the-art performance in zero-shot 3D visual grounding, we acknowledge a few inherent limitations that provide avenues for future research:
\textbf{(1) Computational Overhead and Inference Latency.} 
Our framework relies on a cascading architecture that sequentially integrates multiple foundation models, including CLIP, Gemini, and SAM3. Furthermore, the cross-modal verification and multi-view rendering processes introduce additional computational costs. Consequently, MCM-VG is currently more suited for offline grounding or scenarios without strict real-time constraints. In future work, we aim to explore model distillation or end-to-end unified architectures to accelerate the inference speed.
\textbf{(2) Sensitivity to VLM's Spatial Precision in Extreme Clutter.} 
VLMs inherently possess stronger high-level semantic comprehension than dense, pixel-level geometric reasoning. Although we have introduced a progressive, three-stage point-prompting pipeline (Sec. \ref{sec:det_vcg}) to compensate for this deficiency and mitigate VLM hallucinations, the VLM might still struggle to predict absolutely precise ``positive/negative'' point coordinates in highly cluttered scenes with numerous visually identical instances (e.g., a pile of heavily stacked identical chairs). This geometric ambiguity can subsequently lead to suboptimal SAM segmentation masks. In the future, integrating foundational models with stronger explicit spatial grounding capabilities could further alleviate this bottleneck.
\textbf{(3) Dependence on View Coverage and 2D Visibility.} 
Our ``project-and-rectify'' fallback mechanism (Instance Rectification Module) explicitly shifts trust to 2D visual priors when 3D segmentation fails. This strategy assumes that the target is sufficiently visible in the provided 2D multi-view frames. As shown in Fig.~\ref{fig:failure}, if the target is \textit{severely occluded} by surrounding environments across all available viewpoints, or if it is \textit{heavily truncated} and only marginally visible at the extreme edges of the camera's field of view, the subsequent 2D-to-3D back-projection may inevitably yield incomplete or highly noisy 3D geometries. Enhancing the robustness of 3D geometry completion under adversarial views remains an important direction.

\section{Details of Instance Rectification Module}
\label{ir}

\subsection{More Ablation Studies}
\subsubsection{Different VLMs.}
In the Instance Rectification (IR) module, VLM serves as the core reasoning engine for 2D spatial grounding. To evaluate the robustness and model-agnostic nature of our progressive point-prompting mechanism, we investigate the impact of employing different state-of-the-art VLMs, as presented in Table~\ref{tab:ablation-vlms}.
As observed, replacing the default Gemini-3-pro with other advanced models like GPT-5 or Qwen3.5-plus still yields highly competitive performance, demonstrating that our carefully designed three-stage verification pipeline is generally applicable and not strictly tied to a single VLM architecture. 
Among them, Gemini-3-pro achieves the optimal performance (62.0\% on Acc@0.25 and 53.6\% on Acc@0.5). We attribute this superiority to its exceptional capability in fine-grained spatial reasoning and strict instruction-following. These traits are particularly crucial for our task, as the VLM must precisely predict the intricate ``two positive, one negative'' point coordinates on raw 2D frames while rigorously rejecting semantic hallucinations during the re-verification stage.
\subsubsection{Different Proposal Generation Strategies.}
To validate the rationality of our point-prompting design in the Instance Rectification (IR) module, we evaluate different strategies for generating 2D spatial proposals, as shown in Table~\ref{tab:ablation-strategies}. We compare our approach against two intuitive baselines: relying entirely on the VLM to predict bounding boxes, and feeding the text directly into the 2D segmentation model.
Relying solely on the VLM (Row 1) yields suboptimal results (58.8\% on Acc@0.25). While VLMs possess profound zero-shot semantic comprehension, they inherently lack dense, pixel-level geometric perception. The bounding boxes directly generated by VLMs are frequently coarse and spatially loose, which severely degrades the quality of the subsequent 2D-to-3D back-projection. 
Conversely, relying solely on the segmentation model (Row 2) also achieves constrained performance (58.4\% on Acc@0.25). Although advanced 2D segmentation models excel at pixel-perfect mask generation, their text-encoders typically struggle with complex, free-form, or open-vocabulary language queries without explicit geometric guidance, often leading to semantic mismatches or completely missing the target.
Our proposed synergistic strategy (Row 3) achieves the best performance. This substantial improvement demonstrates an effective decoupling between semantic comprehension and spatial localization. By employing spatial points as a lightweight and unambiguous intermediate representation, our approach successfully synergizes the semantic superiority of VLMs with the pixel-level geometric precision of segmentation foundation models.

\subsubsection{Different Sources of Proposals.}
To validate the core motivation of our cross-modal design, we investigate the necessity of synergizing 2D and 3D visual priors. Table~\ref{tab:ablation-sources} compares the grounding performance when utilizing different sources of initial proposals. 
% Relying exclusively on 2D visual priors (\textbf{2D only}) yields the lowest performance. Without explicit 3D geometric constraints, directly lifting 2D multi-view segmentations into 3D space inevitably suffers from severe depth ambiguities, cross-view inconsistencies, and projection noise. 
Relying exclusively on 2D visual priors (\textbf{2D only}) yields the lowest performance. This degradation stems from two critical bottlenecks. First, without the spatial guidance of 3D proposals to explicitly filter irrelevant multi-view frames, target-absent viewpoints severely mislead the VLM into generating erroneous or hallucinated point prompts. Second, even with correctly generated 2D masks, without explicit 3D geometric constraints, directly lifting 2D multi-view segmentations into 3D space inevitably suffers from depth ambiguities, cross-view inconsistencies, and projection noise.
Conversely, utilizing \textbf{3D only} proposals (e.g., from an off-the-shelf 3D instance segmenter) establishes a respectable baseline. However, its performance is fundamentally bottlenecked by the limited vocabulary and geometric miss-detections of pre-trained 3D networks, which frequently fail to propose masks for unobserved, long-tail, or novel targets.
Our hybrid approach (\textbf{Ours: 3D+2D}) significantly outperforms both single-modality baselines, achieving peak performance (62.0\% on Acc@0.25). This substantial gain explicitly demonstrates the effectiveness of our design philosophy: the 3D proposals provide geometrically robust instance foundations, while the 2D visual cues (leveraged via our Instance Rectification module) serve as a powerful open-vocabulary rescue mechanism. By seamlessly synergizing 3D spatial stability with 2D semantic richness, MCM-VG successfully overcomes the inherent limitations of each individual modality.

\begin{figure*}
    \centering
    \includegraphics[width=0.97\linewidth]{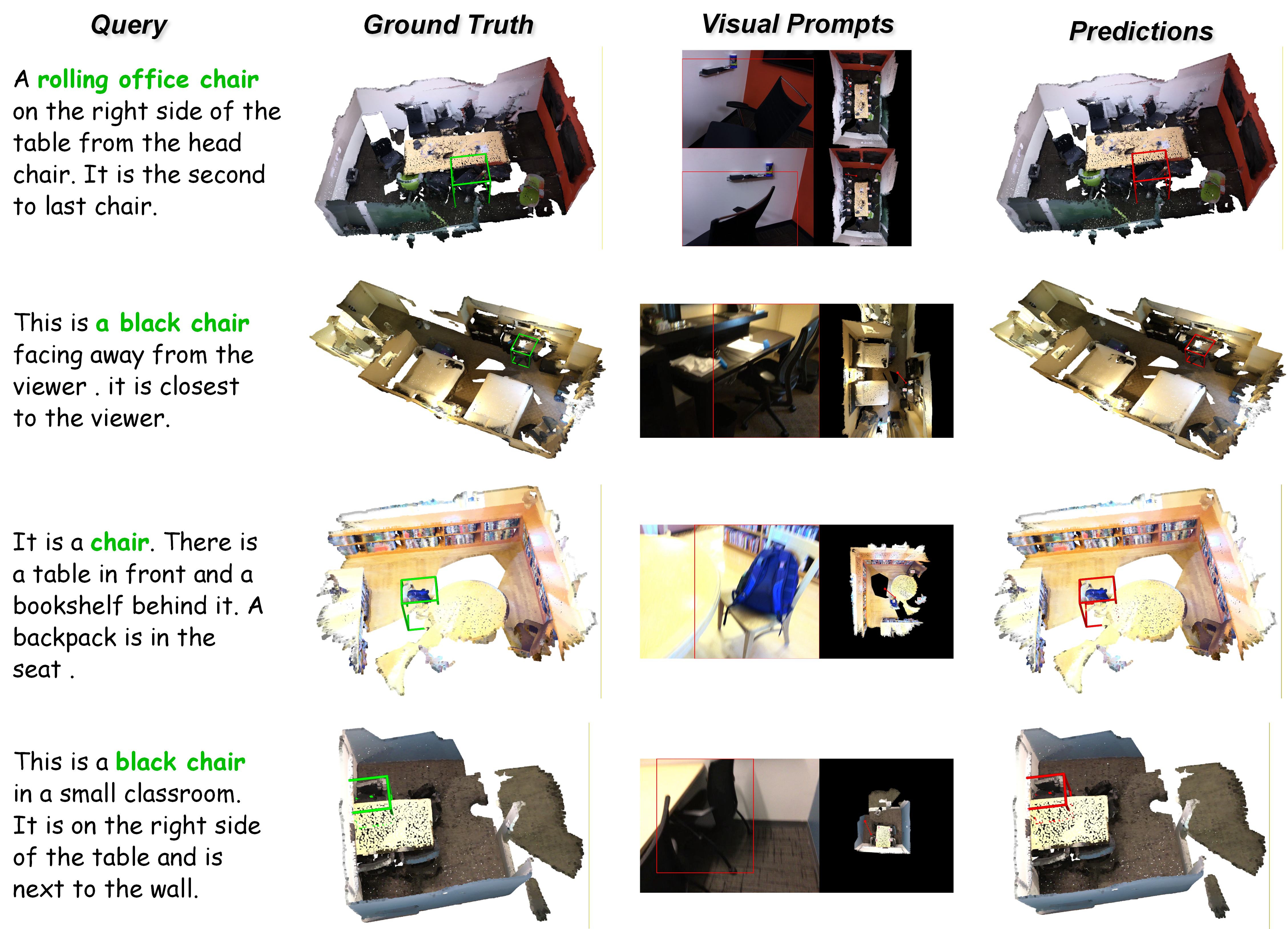}
    \vspace{-0.3cm}

    \caption{More visualization results of 3D visual grounding on the ScanRefer~\cite{chen2020scanrefer} dataset.}
    
    \label{fig:more_vis}
    \vspace{-0.1cm}
\end{figure*}

\begin{table}[t]
    \caption{Ablation study on the Viewpoint Distillation (VD) module. The first row substitutes our VD module with the sequential reasoning strategy from SeqVLM~\cite{lin2025seqvlm}, while both utilize the identical 3D/2D proposals generated by our preceding modules.}
    \vspace{-0.1cm}
    \centering
    \setlength{\tabcolsep}{5.5pt} 
    \renewcommand{\arraystretch}{1.3}
    \begin{tabular}{l | c c c c} 
    \toprule
    \textbf{Strategy} & \textbf{Time (s)} & \textbf{Cost} & \textbf{Input Img.} & \textbf{Acc$@$0.5} \\
    \midrule
    SeqVLM-style & \textbf{27.75} & 17.55k & 5 & 50.0 \\
    \rowcolor{gray!20}
    \textbf{VD (Ours)} & 38.05 & \textbf{15.08k} & \textbf{1.5 ($\times$2)} & \textbf{53.6} \\
    \bottomrule
    \end{tabular}
    \vspace{-0.3cm}
    \label{tab:ablation_vd}
\end{table}

\subsection{Details of Visual Cue Generation}
\label{sec:det_vcg}
In the Instance Rectification (IR) module, robust 2D visual cues are essential for rescuing unlocalized targets. To effectively bridge the gap between open-vocabulary text queries and pixel-level 2D segmentation, we leverage the profound zero-shot reasoning capabilities of a Large Vision-Language Model (i.e., Gemini~\cite{team2023gemini}) via a progressive, three-stage point-prompting pipeline. The prompts are shown in \ref{sec:vcg} and the detailed process is elaborated as follows:

\textbf{Stage 1: Contextual Anchor Filtering and Annotation.} 
We first prompt the VLM to evaluate the raw video frames and perform logical reasoning to eliminate frames irrelevant to the target. Concurrently, we conduct an anchor-related check to retain keyframes that contain the contextual anchor objects parsed from the query. For these retained keyframes, we extract the spatial coordinates of the anchors and explicitly overlay their bounding boxes onto the frames. This step establishes crucial geometric context and reference points for the VLM.
    
\textbf{Stage 2: Target-Aware Point Prompting.} 
Building upon the anchor-annotated frames (or raw frames if no anchors are present), we prompt the VLM to pinpoint the exact location of the target object. To enhance the robustness of the subsequent segmentation, we instruct the VLM to generate a ``two positive, one negative'' point set: two points accurately located on the target body (positive) and one point located on the background or distracting objects (negative). Along with reasoning-based frame filtering, the coordinates of these points are recorded, and we render intermediate visualization images with these points explicitly plotted for the next stage.
    
\textbf{Stage 3: Semantic Verification and 2D Segmentation.} 
Given the inherent hallucination risks in VLM point generation, we introduce a stringent visual re-verification step. We feed the intermediate point-visualized images back into the VLM, querying it to independently re-identify the object category at the marked locations. If the predicted category contradicts the intended target semantics, the frame is immediately discarded to prevent error propagation. For the successfully verified frames, the extracted positive and negative coordinates are directly employed as explicit geometric prompts for an advanced 2D foundation model (i.e., SAM3~\cite{carion2025sam}). Guided by these precise point prompts, SAM generates highly accurate 2D bounding boxes and masks, which subsequently serve as the definitive 2D visual priors for 3D back-projection.

Through this coarse-to-fine visual cue generation mechanism, we ensure that the VLM's semantic comprehensions are accurately and reliably grounded into high-quality 2D geometric representations, while effectively filtering out noise and hallucinations.

\section{Analysis of Viewpoint Distillation Module}
\label{vd}
To explicitly validate the effectiveness of our Viewpoint Distillation (VD) module and the final target disambiguation strategy, we conduct an ablation study isolating the final reasoning phase. As detailed in Table~\ref{tab:ablation_vd}, we compare our proposed RGB-BEV joint reasoning mechanism against the sequential reasoning strategy from SeqVLM~\cite{lin2025seqvlm}. Both variants build upon the identical robust 3D/2D proposals generated by our preceding Semantic Alignment and Instance Rectification modules. Note that the metrics for \textbf{Time} (seconds), \textbf{Cost} (total tokens), and \textbf{Input Images} are averaged per API response cycle.

When substituting our VD module with the SeqVLM reasoning approach (Row 1), the model processes an average of 5 unclustered frames. While this yields a baseline accuracy of 50.0\% on Acc@0.5, the massive visual redundancy leads to a higher token consumption (17.55k). By integrating our VD module (Row 2), the number of input images is drastically reduced to an average of 1.5 pairs (i.e., 1.5 RGB frames paired with 1.5 BEV maps). This reduction demonstrates the power of our spatial clustering, which prunes redundant viewpoints and significantly drops the overall token \textbf{Cost} to 15.08k.

Furthermore, while our method entails a slightly higher average inference \textbf{Time} (38.05s vs. 27.75s), this latency is fundamentally attributed to the complex cross-modal spatial reasoning required to jointly analyze local representative RGB frames and holistic BEV maps. Importantly, this joint 2D-3D reasoning mechanism provides the VLM with unambiguous spatial referentiality, driving a precise 3.6\% absolute improvement in final grounding accuracy (\textbf{Acc@0.5} increases from 50.0\% to 53.6\%). By isolating the reasoning stage, this experiment concretely proves that our distilled multiple-choice formulation achieves a highly worthwhile trade-off: trading a marginal increase in API inference time for minimized token cost and superior spatial accuracy.

% \begin{table}[t]
%     \caption{Comparison of the Final VLM Reasoning.}
%     \vspace{-0.25cm}
%     \centering
%     \setlength{\tabcolsep}{4.0pt} 
%     \renewcommand{\arraystretch}{1.3}
%     \begin{tabular}{l | c c c c} 
%     \toprule
%     \textbf{Method} & \textbf{Time} & \textbf{Cost} & \textbf{Input Images} & \textbf{Acc$@$0.5} \\
%     \midrule
%     w/o VD (SeqVLM-style) & \textbf{27.75} & 17.55k & 5 & 50.0 \\
%     \textbf{w/ VD (Ours full)} & 38.05 & \textbf{15.08k} & \textbf{1.5 (×2)} & \textbf{53.6} \\
%     \bottomrule
%     \end{tabular}
%     \vspace{-0.2cm}
%     \label{tab:ablation_vlm}
% \end{table}

\section{More Visualization Results}
\label{vis}
We present additional qualitative visualization results of our proposed framework in Fig.~\ref{fig:more_vis}, along with the corresponding visual prompt pairs fed into the VLM.

% \clearpage

\section{Prompts of MCM-VG}
To ensure full transparency and facilitate the reproducibility of our work, this section comprehensively details the exact textual prompts employed across the various modules of  MCM-VG.

\label{prompt}
\subsection{Prompt for Top-10 Category Parsing in Semantic Alignment Module}
% \textbf{\textit{Top-10 Category Parsing:}}

\begin{tcolorbox}[
    colback=gray!3,
    colframe=black!25,
    boxrule=0.25pt,
    arc=2pt,
    left=5pt,
    right=5pt,
    top=5pt,
    bottom=5pt,
    width=\linewidth,
    breakable
]
\small
% \textcolor{teal}{\textit{\textbf{Top-10 Category Parsing:}}}

You are given a query sentence describing a target object in a 3D scene and a predefined Object category list. 
Your task is to identify the target object category and select the top 10 most relevant object categories from the provided Object category list that best match the target object described in the query.

Instructions:\\
1. Target category\\
    A. Extract the single most accurate target object category from the query.\\
    B. The target category must appear exactly in the Object category list.\\
2. Spatial references\\
    A. Extract object categories that are mentioned only as spatial reference objects in the query (e.g., “to the left of the lamp”).\\
    B. Store them as a list under spatial\_refs.\\
    C. Do not treat spatial reference objects as target objects.\\
3. Top category candidates\\
    A. Select exactly 10 unique categories from the Object category list.\\
    B. Do not include duplicate categories, even if they appear multiple times in the list.\\
    C. Rank categories from highest to lowest relevance based on:\\
        - Semantic similarity to the target object\\
        - Common furniture taxonomy (parent / sibling categories)\\
        - Typical usage and visual similarity\\
    D. Categories must be chosen strictly from the Object category list. Do not invent, rename, or paraphrase categories.\\

Output format (STRICT):\\
1. Output only a valid JSON object.\\
2. Use exactly the following keys and structure:\\
- query (string)\\
- target\_category (string)\\
- spatial\_refs (list of strings)\\
- top\_categories (list of 10 strings, ordered by relevance)\\
3. Do not add explanations, comments, or extra fields.\\

Output JSON Template:\\
\{ \\
 \hspace*{2em} "query": "<original query sentence>",\\
 \hspace*{2em} "target\_category": "<single category from Object category list>",\\
 \hspace*{2em} "spatial\_refs": ["<category\_1>", "<category\_2>", "..."],\\
 \hspace*{2em} "top\_categories": [\\
 \hspace*{4em} \ "<category\_1>",\\
 \hspace*{4em}  "<category\_2>",\\
 \hspace*{4em}  "<category\_3>",\\
 \hspace*{4em}  "<category\_4>",\\
 \hspace*{4em}  "<category\_5>",\\
 \hspace*{4em}  "<category\_6>",\\
 \hspace*{4em}  "<category\_7>",\\
 \hspace*{4em}  "<category\_8>",\\
 \hspace*{4em}  "<category\_9>",\\
 \hspace*{4em}  "<category\_10>"\\
 \hspace*{2em} ]\\
\}\\
Query sentence: 
\textcolor{blue!60!black}{\{}
\textcolor{orange}{query} 
\textcolor{blue!60!black}{\}}

Object category list: 
\textcolor{blue!60!black}{\{}
\textcolor{orange}{obj\_list}
\textcolor{blue!60!black}{\}}

\end{tcolorbox}

% \subsection{prompt for IR Stage 1}
\subsection{Prompt for Visual Cue Generation in Instance Rectification Module}
\label{sec:vcg}
% \textbf{\textit{Stage 1: Contextual Anchor Filtering and Annotation:}}
\begin{tcolorbox}[
    colback=gray!3,
    colframe=black!25,
    boxrule=0.25pt,
    arc=2pt,
    left=5pt,
    right=5pt,
    top=5pt,
    bottom=5pt,
    width=\linewidth,
    breakable
]
\small
\textcolor{teal}{\textit{\textbf{Stage 1: Contextual Anchor Filtering and Annotation:}}}

You are a JSON generator.
Return exactly ONE JSON object and nothing else (no markdown, no explanation).

Constraints:\\
- Presence must be exactly "Yes" or "No".\\
- If Presence == "No": point must be null.\\
- If Presence == "Yes": point must be [x, y] with integers (pixel coordinates of the object center).\\
- confidence must be a number between 0 and 1.\\
- Output must contain ONLY these 4 keys: Presence, point, confidence, Reasoning.\\

Query: \textcolor{blue!60!black}{\{}
\textcolor{orange}{query}
\textcolor{blue!60!black}{\}}

Anchor IDs (may be empty):
\textcolor{blue!60!black}{\{}
\textcolor{orange}{object\_ids\_description}
\textcolor{blue!60!black}{\}}

Output JSON (example structure, fill with real values):\\
\{ \\
 \hspace*{2em} "Presence": "No",\\
 \hspace*{2em} "point": null,\\
 \hspace*{2em} "confidence": 0.0,\\
 \hspace*{2em} "Reasoning": ""\\
\} \\
\end{tcolorbox}

% \subsection{2 positive points + 1 negative point}
% \textbf{\textit{Stage 2: Target-Aware Point Prompting:}}
\begin{tcolorbox}[
    colback=gray!3,
    colframe=black!25,
    boxrule=0.25pt,
    arc=2pt,
    left=5pt,
    right=5pt,
    top=5pt,
    bottom=5pt,
    width=\linewidth,
    breakable
]
\small
\textcolor{teal}{\textit{\textbf{Stage 2: Target-Aware Point Prompting:}}}

You are a precise Visual Grounding engine. \\
Your goal is to identify the TARGET and provide three distinct feedback points to define its spatial extent and boundaries.\\

Constraints:\\
- Presence: Exactly "Yes" or "No".\\
- Points Definition (If Presence == "Yes"): \\
    1. Positive Point 1 [x1, y1]: The GEOMETRIC CENTER of the target.\\
    2. Positive Point 2 [x2, y2]: A corner or boundary point (e.g., top-left) that defines the scale of the target.\\
    3. Negative Point [x3, y3]: A point clearly OUTSIDE the target's boundary, typically on a nearby distracting object or background, used to clarify the target's limits.\\
- Coordinates: Use NORMALIZED COORDINATES [0-1000] (where [0,0] is top-left and [1000,1000] is bottom-right).\\
- Confidence: A float between 0.0 and 1.0.\\
- Reasoning: You MUST first describe the visual appearance and surrounding context of the target to ensure you haven't confused it with similar nearby objects.\\

Query: 
\textcolor{blue!60!black}{\{}
\textcolor{orange}{query} 
\textcolor{blue!60!black}{\}}

Anchor IDs: 
\textcolor{blue!60!black}{\{}
\textcolor{orange}{object\_ids\_description} 
\textcolor{blue!60!black}{\}}

Return exactly ONE JSON object:\\
\{\\
\hspace*{2em}  "Presence": "Yes",\\
\hspace*{2em}  "positive\_points": [[x1, y1], [x2, y2]],\\
\hspace*{2em}  "negative\_points": [[x3, y3]],\\
\hspace*{2em}  "confidence": 0.95,\\
\hspace*{2em}  "Reasoning": "Identified the [Target Name] based on its [Color/Shape]. Point 1 is the center, Point 2 is the top-left corner. The negative point is placed on the [Distractor Object] to distinguish the target from its background."\\
\}\\
\end{tcolorbox}

% \subsection{prompt for point recheck}
% \textbf{\textit{Stage 3: Semantic Verification and 2D Segmentation:}}
\begin{tcolorbox}[
    colback=gray!3,
    colframe=black!25,
    boxrule=0.25pt,
    arc=2pt,
    left=5pt,
    right=5pt,
    top=5pt,
    bottom=5pt,
    width=\linewidth,
    breakable
]
\small
\textcolor{teal}{\textit{\textbf{Stage 3: Semantic Verification and 2D Segmentation:}}}

Role: You are a visual reasoning assistant specializing in object verification and spatial relationships.
Task: Analyze the provided image and determine if the area marked by the green dots matches the user's query description.\\
Query: 
\textcolor{blue!60!black}{\{}
\textcolor{orange}{query} 
\textcolor{blue!60!black}{\}}

Anchor IDs: 
\textcolor{blue!60!black}{\{}
\textcolor{orange}{object\_ids\_description} 
\textcolor{blue!60!black}{\}}

Instructions: \\
- Identify the Surface: Determine what specific object or surface the green dots are physically touching (e.g., wall, floor, specific furniture).\\
- Evaluate Context: Check if the location matches the spatial clues in the query (e.g., "by the desk").\\
- Verify Object Type: Compare the material and function of the target object to the query (e.g., "wooden" vs. "plaster").\\
- Logical Consistency: Even if the location is correct, if the object type is missing or different, it is a mismatch.\\
Output Requirements: Provide a brief reasoning and a final JSON response.\\
Example:\\
\{\\
\hspace*{2em}   "query\_match": false,\\
\hspace*{2em}   "target\_object": "wall",\\
\hspace*{2em}   "reasoning": "The green dots are placed on a plain white wall surface. While the location is adjacent to a desk/monitor area, there is no small wooden bookshelf visible at the coordinates indicated by the dots."\\
\}\\
\end{tcolorbox}

\subsection{Prompt for Final Reasoning in Viewpoint Distillation Module}
\begin{tcolorbox}[
    colback=gray!3,
    colframe=black!25,
    boxrule=0.25pt,
    arc=2pt,
    left=5pt,
    right=5pt,
    top=5pt,
    bottom=5pt,
    width=\linewidth,
    breakable
]
\small
% \textcolor{teal}{\textit{\textbf{Final Reasoning:}}}

Imagine you are in a room and you are asked to find one object.\\
Given a series of images and a query describing a specific object in the room, you need to analyze the images, and find an image that best fits the query.\\
Please note:\\
1. Each input image is a composite:\\
  - The left side displays perspective camera views from a sequence, where the target object is highlighted by a red rectangle.\\
  - The right side displays the BEV (Bird's-Eye View) map of the room, which represents the top-down spatial layout of the room.\\
2. Understanding the BEV Map: The red dots and arrows in the BEV map represent the camera's spatial position and its viewing direction (orientation) at the moment the left image was captured. \\
3. Spatial Correspondence: The red marker or arrow in the BEV map on the right corresponds to the specific object shown in the red boxes on the left.\\
4. Your Task: You must combine the visual appearance (from the left sub-images) with the spatial context (from the right BEV map, e.g., location in the room, proximity to walls or other furniture) to make your selection.\\

Your response should be in the following format, and it should not include code block markers such as ``` ***.

\{\\
  "process": "Explain the process of how you identified the room's features and located the target object",\\
  "image\_id": 1 \# Replace with the actual index based on the input order of images, starting from 0. \\
\}\\

Here is an example for you.

```
Input: 
Query: Find the black table that is surrounded by four chairs.\\
Here are the images of 3 possible objects.
[image\_0, image\_1, image\_2]\\

Output:\\
\{\\
  "process": "After carefully examining all the input images, I found only the tables in image\_1, image\_2 are black, but only the tables in image\_2 is surrounded by four chairs. So the correct object is the table in image\_2",\\
  "image\_id": 2\\
\}\\

```

Here are some tips:
\# Please follow the format of the example strictly\\
\# If none match perfectly, choose the most plausible one.\\
\# Only if the types of all objects are completely irrelevant with the query, output -1 in the value of image\_id.\\

Query: 
\textcolor{blue!60!black}{\{}
\textcolor{orange}{query} 
\textcolor{blue!60!black}{\}}

Here are the images of 
\textcolor{blue!60!black}{\{}
\textcolor{orange}{n\_images} 
\textcolor{blue!60!black}{\}}
possible objects.

\textcolor{red}{\# IMAGE\_ID\_INVALID\_PROMPT:}

The image\_id 
\textcolor{blue!60!black}{\{}
\textcolor{orange}{images\_id} 
\textcolor{blue!60!black}{\}}
you selected does not exist. Did you perhaps see it incorrectly? Please reconsider and select another image. Remember to reply using JSON format with the two keys "process", "image\_id" as required before.

\textcolor{red}{\# WRONG\_FORMAT\_PROMPT:}

The answer contains extra characters. Please follow the format of the example strictly.\\

\textcolor{red}{\# REFLECTION\_PROMPT:}\\
Wait. You concluded that no image matches based on strict criteria. 
    However, object descriptions (especially shapes like 'trapezoidal prism') can be subjective or distorted by camera angles.\\
    
    Please RE-EVALUATE the images with a slightly relaxed constraint:\\
    1. Focus more on the **Object Category** (e.g., is it a trash can?) and **Color/Texture**.\\
    2. Focus on the **Spatial Location** (e.g., right of the door).
    3. Be tolerant of minor shape discrepancies.\\
    If there is an image that is a STRONG match for category and location, select it even if the shape isn't perfect.\\
    If you still strictly believe none match, return -1.\\

\end{tcolorbox}

\end{document}